\documentclass{article}

\usepackage{arxiv}
\usepackage[utf8]{inputenc} 
\usepackage[T1]{fontenc}    
\usepackage{hyperref}       
\usepackage{url}            
\usepackage{booktabs}       
\usepackage{amsfonts}       
\usepackage{nicefrac}       
\usepackage{microtype}      

\usepackage{comment}
\usepackage{array}
\usepackage{booktabs}
\usepackage{url}
\usepackage{amssymb}
\usepackage{graphicx}
\usepackage{amsmath}
\usepackage{amsthm}
\usepackage{algorithm}
\usepackage{algorithmic}
\usepackage[caption=false]{subfig}
\usepackage{dsfont}
\usepackage{enumerate}
\usepackage[bottom]{footmisc}

\usepackage{wrapfig}

\newcommand{\MSE}{$\textrm{MSE}_{\textrm{HR}}$}

\title{Spacecraft Collision Avoidance Challenge:\\design and results of a machine learning competition}

\author{
  Thomas Uriot \\
  Advanced Concepts Team \\
  European Space Agency \\
  Noordwijk, The Netherlands \\
  \texttt{uriot.thomas@gmail.com} \\
   \And
  Dario Izzo \\
  Advanced Concepts Team \\
  European Space Agency \\
  Noordwijk, The Netherlands \\
  \texttt{dario.izzo@esa.int} \\
  \And
   Lu{\'i}s F. Sim{\~o}es \\
   ML Analytics \\
   Lisbon, Portugal \\
   \texttt{luis.simoes@mlanalytics.pt} \\
    \And
   Rasit Abay \\
   FuturifAI \\
   Canberra, Australia \\
   \texttt{r.abay@futurifai.com}\\
    \And
   Nils Einecke \\
  Honda Research Institute Europe \\
   Offenbach, Germany \\
   \texttt{nils.einecke@honda-ri.de}\\
    \And
   Sven Rebhan \\
   Honda Research Institute Europe \\
   Offenbach, Germany \\
   \texttt{sven.rebhan@honda-ri.de}\\
   \And
   Jose Martinez-Heras \\
   Solenix GmbH \\
   Darmstadt, Germany \\
   \texttt{jose.martinez@solenix.ch} \\
   \And
   Francesca Letizia \\
   IMS Space Consultancy GmbH \\
   Darmstadt, Germany \\
   \texttt{francesca.letizia@esa.int} \\
   \And
   Jan Siminski \\
   IMS Space Consultancy GmbH \\
   Darmstadt, Germany \\
   \texttt{jan.siminski@esa.int}
   \And
   Klaus Merz \\
   Space Debris Office  \\
   European Space Agency, \\
   Darmstadt, Germany \\
   \texttt{Klaus.Merz@esa.int}\\
}

\begin{document}
\maketitle

\begin{abstract}
Spacecraft collision avoidance procedures have become an essential part of satellite operations. Complex and constantly updated estimates of the collision risk between orbiting objects inform the various operators who can then plan risk mitigation measures. Such measures could be aided by the development of suitable machine learning models predicting, for example, the evolution of the collision risk in time. In an attempt to study this opportunity, the European Space Agency released, in October 2019, a large curated dataset containing information about close approach events, in the form of Conjunction Data Messages (CDMs), collected from 2015 to 2019. This dataset was used in the Spacecraft Collision Avoidance Challenge, a machine learning competition where participants had to build models to predict the final collision risk between orbiting objects. This paper describes the design and results of the competition and discusses the challenges and lessons learned when applying machine learning methods to this problem domain.

\end{abstract}

\newpage

\section{Introduction}

The overcrowding of the Low Earth Orbit (LEO) is a known fact extensively discussed in the scientific literature \cite{liou2008instability, krag2012consideration}. It is estimated that more than 900,000 small debris objects with at least 1cm in radius are currently orbiting uncontrolled in LEO\footnote{Data from \url{https://sdup.esoc.esa.int/discosweb/statistics/}, last access 03/06/2020}, posing a threat to operational satellites \cite{klinkrad2010space}. 
The consequences of an impact between orbiting objects can be dramatic, as the 2009 Iridium-33/Cosmos-2251 collision demonstrated \cite{anselmo2009analysis}. While in the case of impacts with smaller objects a shielding of the satellite may be effective \cite{ryan2013hypervelocity}, any impact of an active satellite with objects that have a cross section larger than 10cm is most likely resulting in its complete destruction. International institutions and agencies grew increasingly concerned over the past decades and contributed to define guidelines to mitigate collision risk and preserve the space environment for future generations \cite{iadc}. As a result, agencies, together with operators and manufacturers, have been assessing a number of approaches and technologies in the attempt to alleviate the issue \cite{walker2004cost, biesbroek2017deorbit, liou2010controlling}. 

\begin{figure}[b]
    \begin{minipage}{.4575\textwidth}
    \centering
    \includegraphics[width=\linewidth]{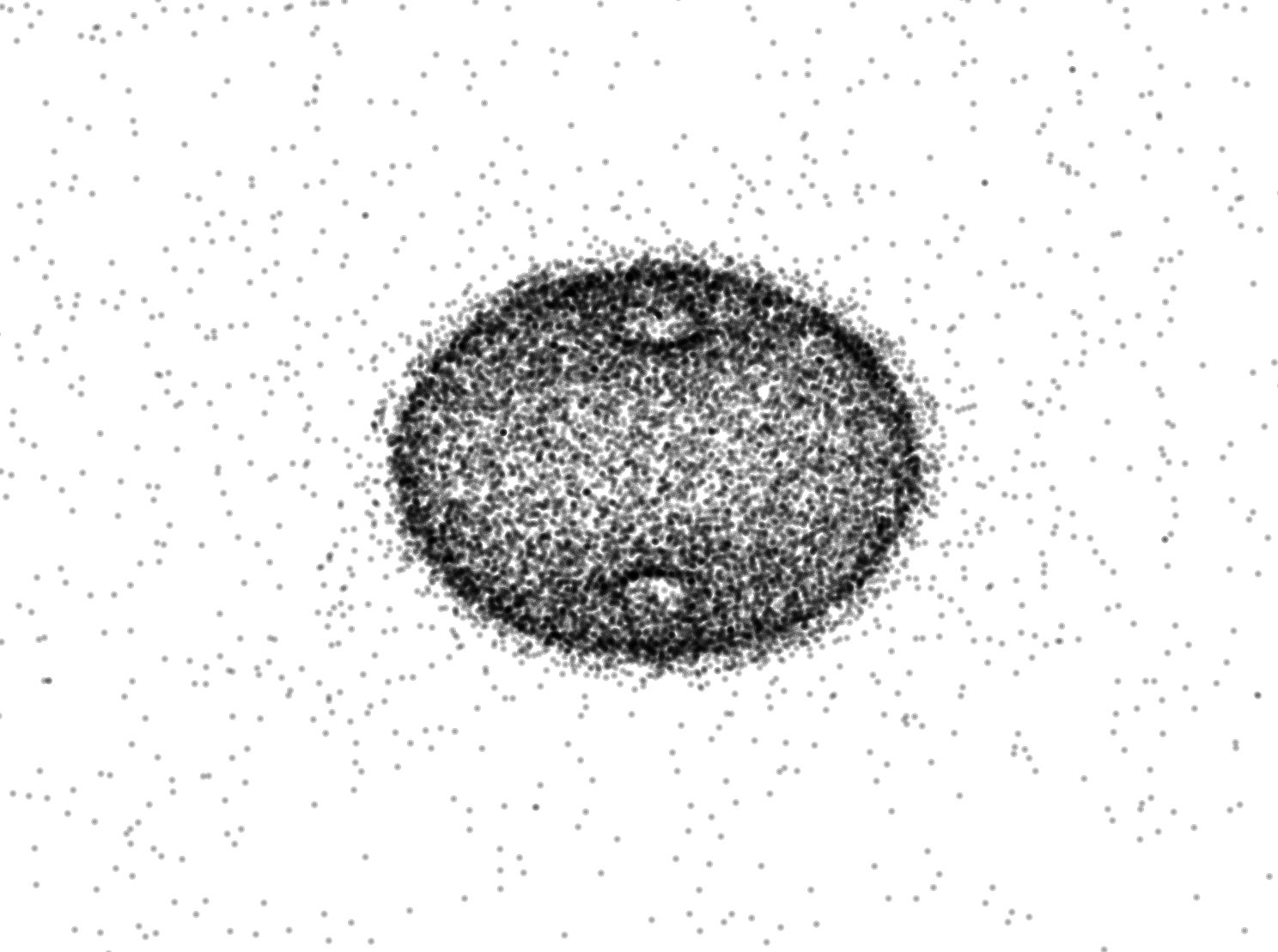}
    \caption{Visualization of the density of Low Earth Orbit orbiting objects as of 2020-May-22 (data from www.space-track.org).}
    \label{fig:leo}
    \end{minipage}
    \hspace*{\fill}
    \begin{minipage}{.51\textwidth}
    \centering
    \includegraphics[width=\linewidth]{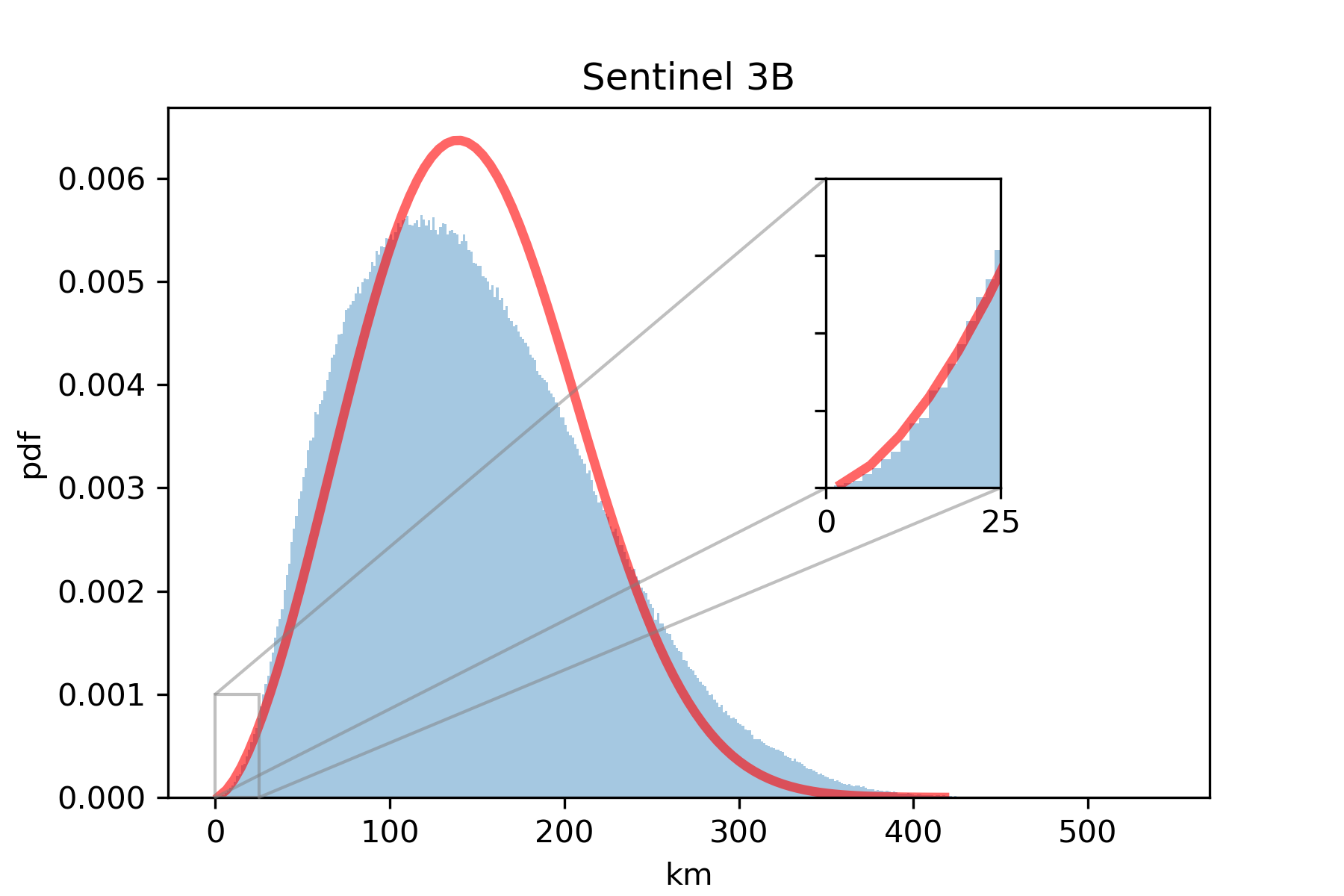}
    \caption{Distribution of the distance between the closest object and Sentinel-3B, and a fitted Weibull distribution (fit skewed to represent the left tail with higher accuracy).}
    \label{fig:weibull}
    \end{minipage}
\end{figure}

Despite all the efforts put in the active control of the debris and of the satellite populations, the situation is still of growing concern today. To visualize just how crowded some areas of the Low Earth Orbit are, we have visualized, as of the 22nd of May 2020, the position of all the 19,084 objects monitored by the radar and optical observations of the US Space Surveillance Network (SSN) and released in the form of NORAD two line elements (TLEs) in Figure~\ref{fig:leo}.
The density of objects at low altitudes appears evident as well as the density drop around the northern and southern polar caps due to the orbital dynamics being dominated by the main orbital perturbations that, in LEO, act mainly on the argument of perigee and on the right ascension of the ascending node \cite{izzo2005effects}. 

To get a first assessment of the risk posed to an active satellite operating, for example, in a Sun-synchronous orbit, we compute at random epochs and within a two-year window, the closest distance of a Sun-synchronous satellite to the LEO population and its distribution. In the case of Sentinel-3B, the Figure~\ref{fig:weibull} shows the result. 
In most of the epochs, the satellite is far from other objects, but in some rare cases, the closest distance approaches values that raise concern. 
A Weibull distribution can be fitted to the obtained data where results from extreme value statistics justify its use to make preliminary inferences on collision probabilities \cite{smirnov2001space}. 
Such inferences are very sensitive to the Weibull distribution parameters and in particular to the behaviour of its tail close to the origin. 

This type of inferences as well as a series of resounding events including the destruction of Fengyun-1C (2007), the Iridium-33/Cosmos-2251 collision (2009), and the Briz-M explosion (2012), convinced most satellite operators to include the possibility of collision avoidance manoeuvres in the routine operation of their satellites \cite{flohrer2015operational}.

In addition, the actual number of active satellites is increasing steadily and the plans for mega-constellations such as Starlink, OneWeb, Project Kuiper and others \cite{Logue2019} makes it likely to significantly increase the active satellite population in the coming decades. It is thus also expected that satellite collision avoidance systems will be increasingly important, and their further improvement, in particular their full automation, a priority in the coming decades \cite{flohrer2019cream}.

\subsection*{The Spacecraft Collision Avoidance Challenge}

In order to advance the research on the automation of preventive collision avoidance manoeuvres, the European Space Agency released a unique real-world dataset containing time series of events representing the evolution of collision risk, related to a number of actively monitored satellites. The dataset was made available to the public, as part of a machine learning challenge dubbed the Collision Avoidance Challenge, hosted on the Kelvins online platform\footnote{Hosted at: \url{https://kelvins.esa.int/}}. The challenge took place over a period of two months, with 96 teams participating and resulting in 862 submissions. It attracted a wide range of backgrounds, from students, to machine learning practitioners and aerospace engineers as well as academic institutions and companies. In this challenge, participants were asked to predict the final risk of collision at time of closest approach between a satellite and a space object, using data cropped at two days to the time of closest approach.

In this paper, we analyse the competition's dataset and results, highlighting issues to be addressed by the scientific community in order to advantageously introduce machine learning in the collision avoidance systems of the future. The paper is structured as follows: In Section~\ref{sec:currentCAS}, we describe the collision avoidance pipeline currently in place at the European Space Agency, introducing important concepts used throughout the paper and crucial to the understanding of the dataset. In Section~\ref{sec:dataset}, we describe the dataset and details on its acquisition. Then, in Section~\ref{sec:design}, we outline the competition design process and discuss some of the decisions made and their consequences. The competition results, the analysis of the received submissions and of the challenges faced when building statistical models of the collision avoidance decision making process are the subject of Section~\ref{sec:results}. In Section~\ref{sec:post_comp_ml} we evaluate how well machine learning models generalize in this problem beyond their training data. Finally, we conclude in Sections~\ref{sec:lessons_learned}--\ref{sec:conclusions} by recording the lessons learned during the competition and suggesting future directions for the research in this area.

\section{Collision Avoidance at ESA}
\label{sec:currentCAS}

A detailed description of the Collision Avoidance process currently implemented at the European Space Agency can be found in previous works \cite{merz2017cola, braun2016operational}.
In this section, we briefly outline several fundamental concepts. 

\begin{figure}
    \centering
    \includegraphics[width=0.6\linewidth]{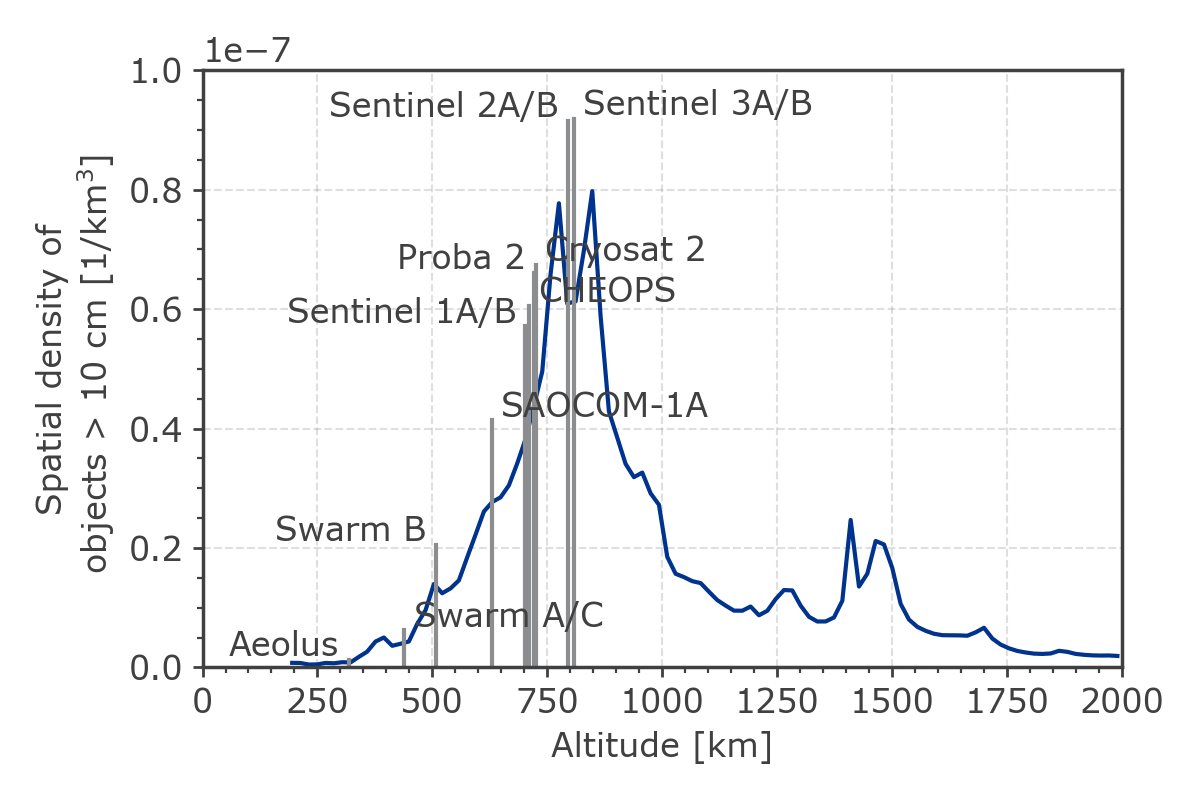}
    \caption{Operational altitudes for the missions in LEO supported by ESA Space Debris Office, and the spatial density of objects with a cross section $>10cm$.}
    \label{fig:supported_missions}
\end{figure}

At the European Space Agency (ESA), the space debris office supports operational collision avoidance activities. Its activities are mainly covering ESA’s missions Aeolus, Cluster II, Cryosat-2, the constellation of Swarm-A/B/C, and the Copernicus Sentinel fleet composed by seven satellites, but also missions of third-party customers. The altitudes of these missions plotted against the background density of orbiting objects as computed by the ESA MASTER8\footnote{Available at \url{https://sdup.esoc.esa.int/master/}} are reported in Figure \ref{fig:supported_missions}.

The main source of information of ESA's collision avoidance process is based on Conjunction Data messages (CDM). These are \emph{ascii} files issued and distributed by the US-based Combined Space Operations Center (CSpOC). 
Each conjunction contains information on one close approach between some monitored space object, the \lq\lq primary satellite\rq\rq\ and a second resident space object, the \lq\lq conjuncting satellite\rq\rq.
The CDMs contain multiple attributes about the approach, such as the identity of the satellite in question, the object type of the potential collider, the time of closest approach (TCA), the positions and velocities of the objects and their associated uncertainties (i.e. covariances), etc. The data contained in the CDM is then processed to obtain risk estimates, applying algorithms such as Alfriend-Akella \cite{alfriendakella}.  

In the days following the first CDM, regular CDM updates are received: over time, the uncertainties of the objects positions become smaller as the knowledge on the close encounter is refined. 
Typically, a time series of CDMs covering one week is released for each unique close approach, with about three CDMs becoming available per day. 
For a given close approach, the last obtained CDM can be assumed to be the best knowledge available about the potential collision and the state of the two objects in question. 
In case the estimated collision risk for a given event is close or above the reaction threshold (e.g. 10$^{-4}$) the Space Debris Office will alarm control teams and start thinking about a potential avoidance manoeuvre a few days prior to the close approach, also meeting with the flight dynamics and mission operations teams.

\section{The database of conjunction events} 
\label{sec:dataset}


\begin{wraptable}{R}{0.4\textwidth}
\centering
\caption{The database of conjunction events at a glance.}
\label{tab:dataset}
\begin{tabular}{ll}\\
\toprule  
\textbf{Characteristics}     &  \textbf{Number}  \\
\midrule
Events  &  15321   \\
High-risk events ($r \geq 10^{-4}$) & 30  \\
High-risk events ($r \geq 10^{-5}$) & 131  \\
High-risk events ($r \geq 10^{-6}$) & 515  \\
\midrule
CDMs & 199082  \\
Average CDMS per event  &  13    \\
Maximum CDMS per event  &  23  \\
Minimum CDMS per event  &  1  \\
\midrule
Attributes & 103 \\
\bottomrule
\end{tabular}
\end{wraptable}

The CDMs collected by ESA's Space Debris Office in the support of collision avoidance operations between 2015 and 2019 was assembled into a database of conjunction events. 
Two initial phases of data preparation were performed. 
First, the database of collected CDMs was queried to consider only events where the theoretical maximum collision probability (i.e. the maximum collision probability obtained by scaling combined target-chaser covariance) was above $10^{-15}$. 
Here, target refers to the ESA satellites while chaser refers to the space debris/object to be avoided. 
In addition, events related to intra-constellation conjunctions (e.g. for the case of the Cluster II) and anomalous entries, as for example cases with null relative velocity between target and chaser, were removed.
Finally, some events may cover a period during which the spacecraft performed a manoeuvre. In these cases, the last estimation of the collision risk surely cannot be predicted from the evolution of the CDM data as the propulsive manoeuvre is not described there. 
These cases were handled by removing all CDM data before the manoeuvre epoch.

A second step in the data preparation was the anonymization of the data. This involved transforming absolute time stamps and position/velocity values in relative values, respectively, in terms of time to TCA and state with respect to the target. The names of the target mission were also removed and a numerical mission identifier was introduced to group similar missions. A random event identifier was assigned to each event. The full list of the attributes extracted from the CDMs and released in the dataset can be found, together with their explanation, on the Kelvins competition website. Here, we briefly describe only a few attributes that are relevant for later discussions:
\begin{itemize}
\item \emph{time\_to\_tca}: time interval between the CDM creation and the time-of-closest approach [days].

\item \emph{c\_object\_type}: type of the object at collision risk with the satellite.

\item \emph{t\_span}: size of the target satellite used by the collision risk computation algorithm [m].

\item \emph{miss\_distance}: relative position between chaser and target.

\item \emph{mission\_id}: identifier of the mission from which the CDMs come from.

\item \emph{risk}: self-computed value at the epoch of each CDM, using the attributes contained in the CDM, as mentioned in Section \ref{sec:currentCAS}. 
\end{itemize}
Table~\ref{tab:dataset} gives an overview of the resulting database, reporting the number of entries (i.e. CDMs) and of unique close approach events. The risk computed from the last available CDM is denoted by $r$.

\section{Competition Design}
\label{sec:design}

The database of conjunction events constitutes an important historical record of risky conjunction events that happened in LEO and opens up the opportunity to test the use of machine learning approaches in the collision avoidance process.
The decision on whether to perform an avoidance manoeuvre is based on the best knowledge one has of the associated collision risk at the time when said manoeuvre cannot be further delayed, i.e. the risk reported in the latest CDM available. 
Such a decision would clearly benefit from a forecast of the collision risk, allowing to take into account also its past evolution and projected trend. When designing the Spacecraft Collision Avoidance Challenge, it was thus natural to start from a forecasting standpoint, seeking an answer to the question: can a machine learning model forecast the collision risk evolution from available CDMs? 

Such a forecast could assist the decision of whether or not to perform an avoidance manoeuvre, by providing a better estimate of the future collision risk, before further CDMs are released. 
Forecasting competitions\cite{hyndman2020brief} are widely recognized as an effective mean to find good predictive models and solutions for a given problem. To design such competitions successfully, requires to find a good balance between the desire to create an interesting and fair machine learning challenge, to motivate and involve the large community of data scientists worldwide, and to fulfil the objective of furthering the current understanding by answering a meaningful scientific question \cite{kisantal2019satellite}.

\begin{wrapfigure}{R}{0.5\textwidth}
    \centering
    \includegraphics[width=\linewidth]{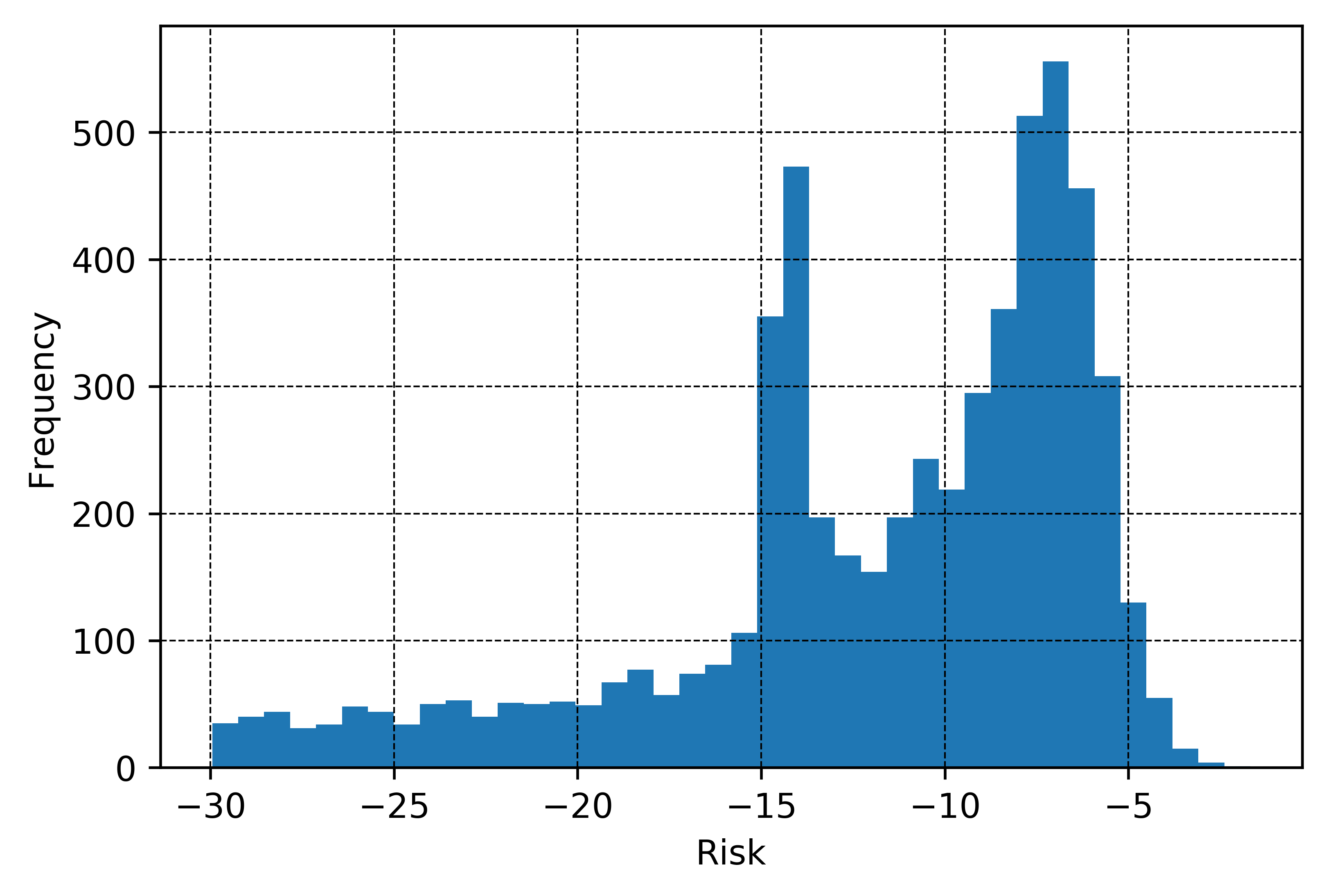}
    \caption{Histogram of the latest known risk value for the whole dataset (training and testing sets). Note that there are 9505 events with a final risk value of -30 or lower, that are not displayed on this figure.}
    \label{fig:risk_histo}
\end{wrapfigure}

Designing a competition to forecast $r$ from the database of conjunction events presented a few challenges. First, the distribution of the final risk $r$ associated to all the conjunction events contained in the database (see Figure~\ref{fig:risk_histo}) is highly skewed, revealing how most events result, eventually, in a negligible risk. 
Intuitively, the effect is due to the uncertainties being reduced as the objects get closer, which in most cases result in close approaches where a safe distance is kept between the orbiting objects. Furthermore, events which already required an avoidance manoeuvre are removed from the data, thus reducing the number of high-risk events.
This is particularly troublesome as the interesting events, the ones that are to be forecasted accurately, are the few ones where the final risk is significant. 
Second, there is a great heterogeneity in the various time series associated to different events, not only in terms of the number of available CDMs, but also of the actual \emph{time\_to\_tca} at which CDMs are available, and most importantly of the \emph{time\_to\_tca} of the last available CDM which defines the variable $r$ to be predicted. The test and training set and the competition metric were thus designed in an attempt to alleviate these issues.

\subsection{Definition of high risk events}

Many mission operators in LEO use $10^{-4}$ as a risk threshold for implementing an avoidance manoeuvre. Over time, this value has started to be applied by default. However, the selection of a suitable reaction threshold for a given mission depends on many different parameters (e.g. chaser's size, target satellite) and its selection can be driven by considerations on the risk reduction that an operator wants to achieve \cite{merz2019risk}. For this reason, ESA missions in LEO adopt reaction thresholds ranging between $10^{-5}$ and $10^{-4}$. Events start to be monitored and highlighted once the collision risk is larger than a notification threshold, which is typically set one order of magnitude lower than the reaction threshold. Note that in the remainder of this paper, the $\log_{10}$ of the risk value is used frequently, so that for example $\log_{10} r \geq -6$ defines high risk events. We will thus often omit to write $\log$ and simply refer to $r \geq -6$.  
For the purpose of the competition, a single notification threshold was used for all the missions and its value was set at $10^{-6}$. 
The threshold value was chosen to have a higher number of high-risk events while keeping its value close to the more frequently used operational value of $10^{-5}$. 
Figure \ref{fig:risk_histo} shows the risk computed from the last available CDM for all the close approach events in the database, revealing a sharp increase around the risk value of -6. In particular, there are 30 events with $r > -4$, 131 events with $r > -5$ and 515 events with $r > -6$ (see Table \ref{tab:dataset}).

\subsection{Test and training sets}
\label{sec:test_train}

While it was a priority to release to the public the raw database of conjunction events, and thus provide the community with an unbiased set of information to learn from, the various models produced during the competition were tested mainly on predictions of events deemed as particularly meaningful. As a consequence, while the training and test sets originated from a split of the original database, they were not randomly sampled from it. Events corresponding to useful operational cases were favoured to appear in test set. 

In particular, for some events, the latest available CDM is days away from the (known) time to closest approach which makes its prediction (also if correct) not a good proxy for the risk at the time to closest approach (TCA). 
Furthermore, potential avoidance manoeuvres are planned at least 2 days prior to closest approach, thus events that contain several CDMs at least 2 days prior to TCA are more interesting.
Overall, three constraints were put on events to be eligible in the test set:
\begin{enumerate}[i]
    \item The event had to contain at least 2 CDMs, one to learn from and one to use as the target.
    \item The last CDM released for the event had to be within 1 day (\emph{time\_to\_tca} $<1$) of the TCA. 
    \item The first CDM released for the event had to be at least 2 days before TCA (\emph{time\_to\_tca} $\geq 2$) and all the CDMs that were within two days from TCA  (\emph{time\_to\_tca} $<2$) were removed. 
\end{enumerate}

\begin{wrapfigure}{R}{0.5\textwidth}
    \centering
    \includegraphics[width=\linewidth]{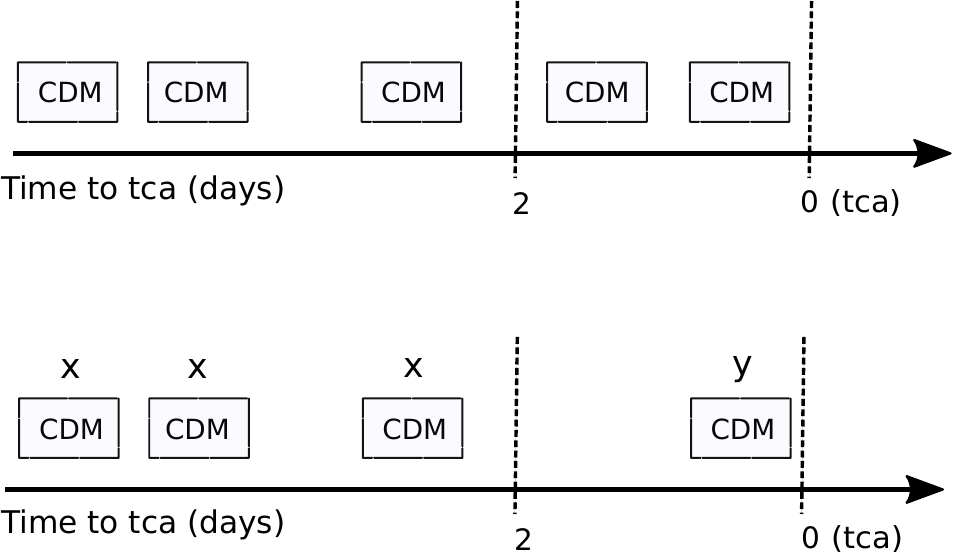}
    \caption{Diagram depicting the raw CDMs time series for one event (top), and the same series in case it is selected for the test set (bottom): only the CDMs prior to 2 days to TCA are made available (labeled as $x$) and the latest CDM is used as the target (labeled as $y$).}
    \label{fig:cdm_diagram}
\end{wrapfigure}

An example of an event which meets the requirements is depicted in Figure \ref{fig:cdm_diagram}. Note that by only allowing events that satisfy the aforementioned requirements in the test set, the number of high-risk events is considerably diminished. After enforcing the three requirements described above, only 216 high-risk events (out of the 515) are eligible for the test set. Note that the remaining 299 high-risk events are left in the training set without being necessarily representative of test events.

Given the unbalanced nature of the dataset, and the small number of high-risk events eligible for the test set, it was decided to put most of the eligible events into the test set. Specifically, 150 of the eligible high-risk events were put in the test set and 66 in the training set. To alleviate the risk of probing directly the test set and thus overfitting, the number of submissions per team was limited to two per day during the first month of the competition, and to a single submission per day during the second month. 

\subsection{Competition Metric}
\label{sec:metric}


In this section, we introduce the metric that was used to rank the participants and discuss its advantages and drawbacks. Several criteria were used to design a metric  that could be fair and reward models of interest for operational purposes.
The Spacecraft Collision Avoidance Challenge had two main objectives: (i) the correct classification of events into high risk and low risk events; (ii) the prediction of the risk value for high risk events. 
In other words, whenever an event belongs to the low risk class, the exact risk value is of no importance, on the other hand, if an event belongs to the high risk class,  its exact value is of interest. 
Furthermore, since in the context of collision avoidance false negatives are a lot more disastrous than false positives, their occurrences should be penalized more.
Finally, this is a highly unbalanced problem, where the proportion of low-risk is much higher than high-risk events.

The final metric used takes these requirements into account and summarizes them into one overall value, in order to rank competitors. Eventually, the Spacecraft Collision Avoidance Challenge metric included both a classification part and a regression part. Denoting the final risk by $r$ and the corresponding prediction by $\hat{r}$, the metric is defined as:

\begin{equation}
\label{eq:metric}
    L(\hat{r}) =  \frac{1}{F_2}\textrm{MSE}_{\textrm{HR}}(r,\hat{r}),
\end{equation}
where $F_2$ is computed over the whole test set, using two classes, (high final risk: $r \geq -6$, low final risk: $r < -6$) and the $\textrm{MSE}_{\textrm{HR}}(r,\cdot)$ is only computed for the high risk events.  More formally, we have 

\begin{equation*}
\textrm{MSE}_{\textrm{HR}}(r,\hat{r}) =   \frac{1}{N^*}\sum\limits_{i=1}^N \mathds{1}_{i} (r_i - \hat{r}_i)^2,
\end{equation*}
where $N$ is the total number of events, $N^* = \sum\limits_{i=1}^N \mathds{1}_{i}$ is the number of high risk events, $r_i$ and $\hat{r}_i$ are the true risk and predicted risk for the $i^{\textrm{th}}$ event and

\[\mathds{1}_{i}=
\begin{cases}
1 \hspace{0.4cm} \textrm{if \hspace{0.1cm} $r_i \geq -6$},\\
0 \hspace{0.4cm} \textrm{otherwise}.
\end{cases} \]
Finally, the $F$ score is defined as

\[F_{\beta} =  (1+\beta^2)\frac{p \times q}{(\beta^2 \times p)+q},\]
where $\beta$ essentially controls the trade-off between precision and recall denoted by $p$ and $q$, respectively. A higher value of $\beta$ means that recall has more weight than precision and, thus, there is more emphasis on false negatives. In order to penalize false negatives more, it was set $\beta=2$. 

While the metric indeed encourages participants to have a higher $F_2$ score and a lower mean squared error, it introduces many layers of subjectivity. This is because the metric contains multiple sub-objectives that are all combined into one meta-objective. In the denominator, the $F_2$ score is already an implicit multi-objective metric, where precision and recall are to be maximized to 1. Indeed, there is a trade-off between precision and recall, which is being controlled by $\beta$. In the numerator, the mean squared error penalizes erroneous predictions for high-risk events. The squaring is justified by the desire to penalize large errors more.

In the metric defined in Eq.~(\ref{eq:metric}), $F_2$ acts as a scaling factor to the $\textrm{MSE}_{\textrm{HR}}$, where $F_2$ takes values in $[0,1]$ and the $\textrm{MSE}_{\textrm{HR}}$ in $\mathrm{R}^+$, which means that the metric will be largely dominated by the $\textrm{MSE}_{\textrm{HR}}$ in the numerator. Nonetheless, as reported in Section 5, even the highest ranked models achieved a relatively small $\textrm{MSE}_{\textrm{HR}}$ and thus the $F_2$ scaling factor is appropriate.

In conclusion, several objectives were combined into one metric, which introduces some level of complexity and subjectivity. An alternative to the metric used in Eq.~(\ref{eq:metric}) would be to have a simple weighted average of each sub-metric ($F_2$ and $\textrm{MSE}_{\textrm{HR}}$). This scoring scheme is routinely used in public benchmarks such as the popular GLUE \cite{wang2019glue} score used in natural language processing, and it presents similar problems in the choice of the weights which act as scaling factors.

It is important to note that according to the Eq.~(\ref{eq:metric}), as soon as an event is predicted to be low-risk ($\hat r < -6$) the optimal prediction to assign to the event is $\hat r = -6 - \epsilon$, where $\epsilon > 0$. 
In this way, in the case of a false negative, we are at least minimizing the $\textrm{MSE}_{\textrm{HR}}$ and in the case of a true negative it does not matter what the actual value is, as long as $\hat r < -6$. 
As a consequence, all risk predictions can be clipped at a value slightly lower than $10^{-6}$ to improve the overall score (or at least produce an equivalent score). In the rest of this paper, we make use of this clipping and the scores of the various teams are reported after the clipping has been applied, using $\epsilon = 0.001$.

\subsection{Baselines}
\label{sec:baseline}

In order to have a sense of how good a proposed solution is, it is useful to introduce baseline solutions. In the case of the Spacecraft Collision Avoidance Challenge, two simple ideas can be used to build such baselines. 
Let us denote $\hat{r}_i$ and ${r}_{{-2}_i}$ as the predicted risk and the latest known risk for the $i^{\textrm{th}}$ event (the subscript $-2$ reminds us that the latest known risk for a close approach event is associated to a CDM released at least two days before TCA, as shown in Figure \ref{fig:cdm_diagram}). The first baseline solution, called Constant Risk Prediction (CRP baseline) is then defined as:
$$
\hat{r}_i = -5
$$
and has an overall score of $L= 2.5$. It constantly predicts the same value for the risk and was highlighted during the competition as a persistent entry in the leaderboard. Out of the 97 teams 38 managed to produce a better model.

One of the simplest approaches in time series prediction is the naive forecast \cite{Hynsman2018}, i.e. forecast with the last known observation. This is known to be actually optimal for random walk data and works well on economic and financial time series.
Based on this fact, a second baseline solution, called Latest Risk Prediction baseline (LRP baseline) is defined as the clipped naive forecast:

\[\hat{r}_i=
\begin{cases}
{r}_{{-2}_i} \hspace{1cm} \textrm{if \hspace{0.1cm} ${r}_{{-2}_i} \geq -6$},\\
-6.001 \hspace{0.2cm} \textrm{otherwise}.
\end{cases} \]
and has a score of $L= 0.694$ when evaluated on the complete test set. Out of the 97 teams 12 managed to submit better solutions. Quite a few different teams found and made use of this baseline (or equivalent variants) in their submissions.
The score achieved by the LRP is also reported in Table \ref{tab:ranking} and is plotted as an horizontal line in Figure \ref{fig:score_evolution}, alongside the proposed solutions from the top 10 teams.
The LRP is of interest in this competition, as in any forecasting competition, because it provides a simple and yet surprisingly effective benchmark to improve upon.

\begin{wrapfigure}{R}{0.5\textwidth}
    \includegraphics[width=0.98\linewidth]{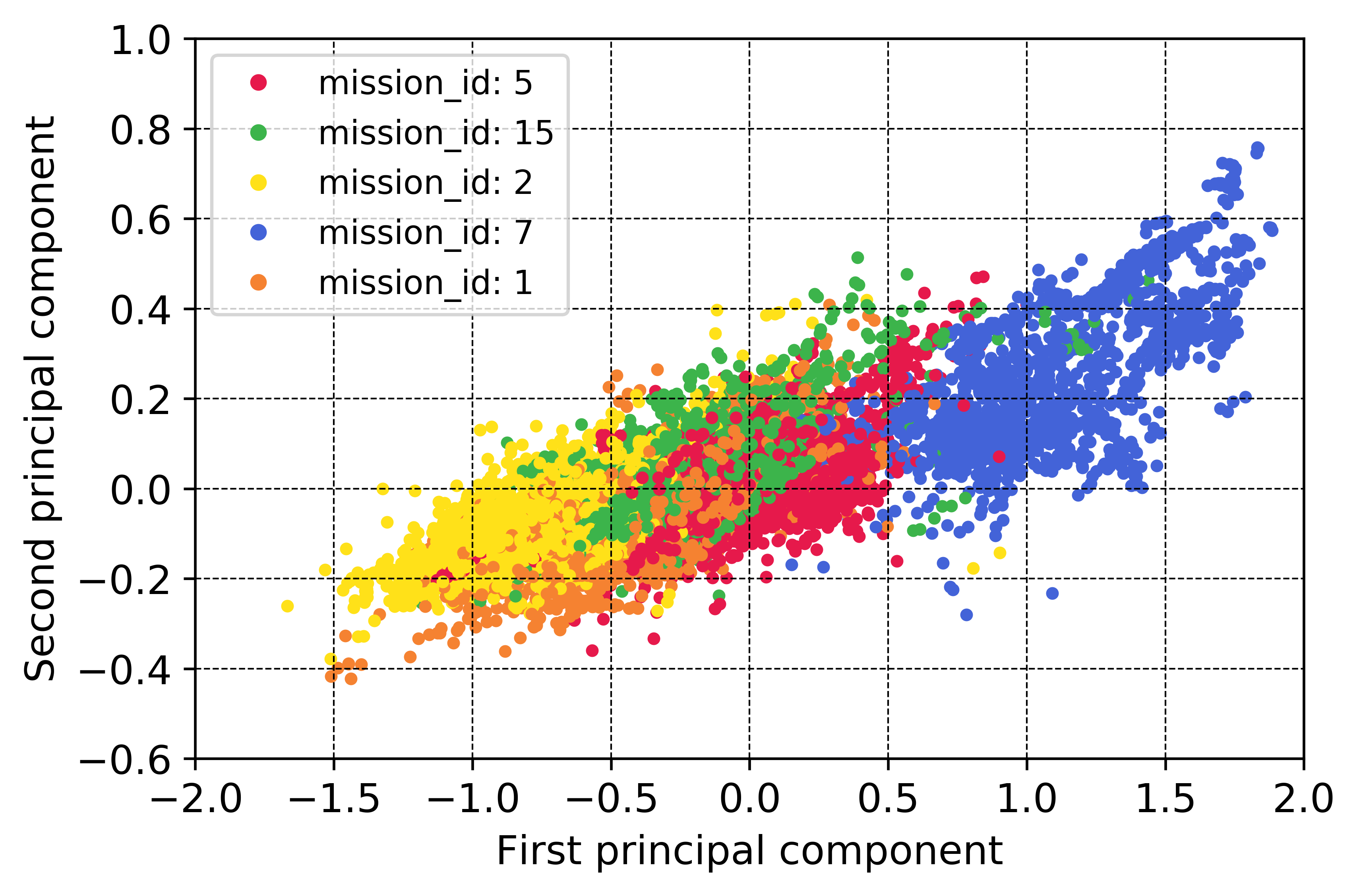}
    \caption{Projection of the original CDMs, from the test set, onto the first two principal components, colored according to \emph{mission\_id}.}
    \label{fig:pca_plot}
\end{wrapfigure}

\newpage

\subsection{Data split}
\label{sec:data_split}

In this section, the split of the original dataset into training and test sets is discussed. Firstly, Principal Components Analysis (PCA) is applied to the data and shows that the attributes depend on the mission identifier. In other words, attributes recorded during different missions are not from the same distribution, making it hard to generalize from one mission to another. Next, we study how different splits of the test data influence the score on the leaderboard (evaluated on a portion of the test set) and in the final ranking (evaluated on the full test set), using the LRP baseline solution. 

In Figure \ref{fig:pca_plot}, the PCA projection of the original data is shown, by only keeping the first two principal components. While the first two principal components only account for $20\%$ of the total variance, the projected data can still be distinguished and roughly clustered by \emph{mission\_id}, in particular \emph{mission\_id: 7} and \emph{mission\_id: 2}. This implies that, not surprisingly, the attributes from the CDMs do not come from the same distribution, making it potentially difficult to generalize from one mission to the other. Indeed, each \emph{mission\_id} refers to a different satellite, orbiting at different altitudes, in regions of space with varying space debris density, as shown in Figure \ref{fig:supported_missions}.

\begin{figure}[t]
    \begin{minipage}[t]{.48\textwidth}
    \centering
    \includegraphics[width=\linewidth]{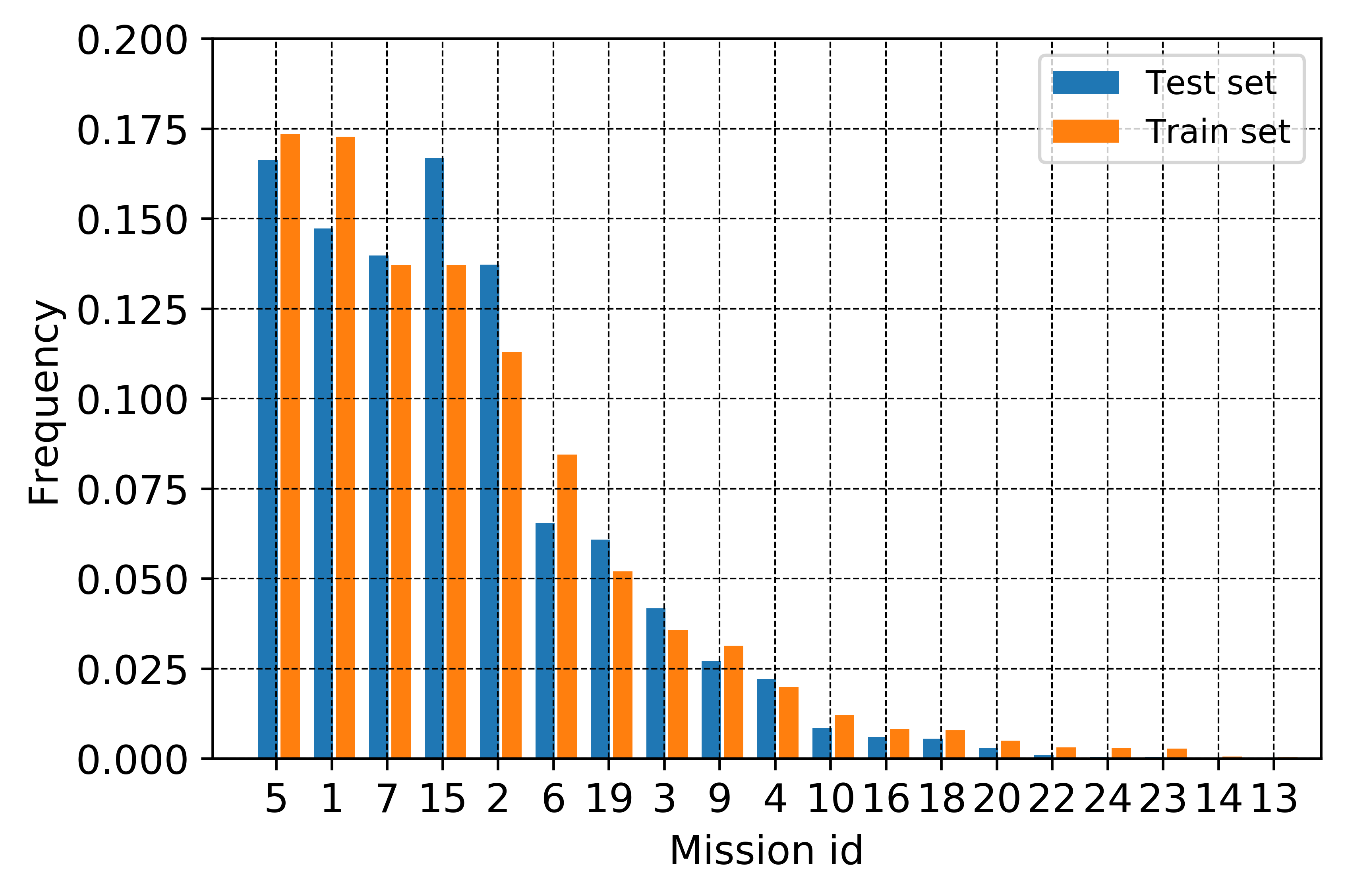}
    \caption{Distribution of the mission type for the test and training sets, for low-risk events.}
    \label{fig:mission_all}
    \end{minipage}
    \hspace*{\fill}
    \begin{minipage}[t]{.48\textwidth}
    \includegraphics[width=\linewidth]{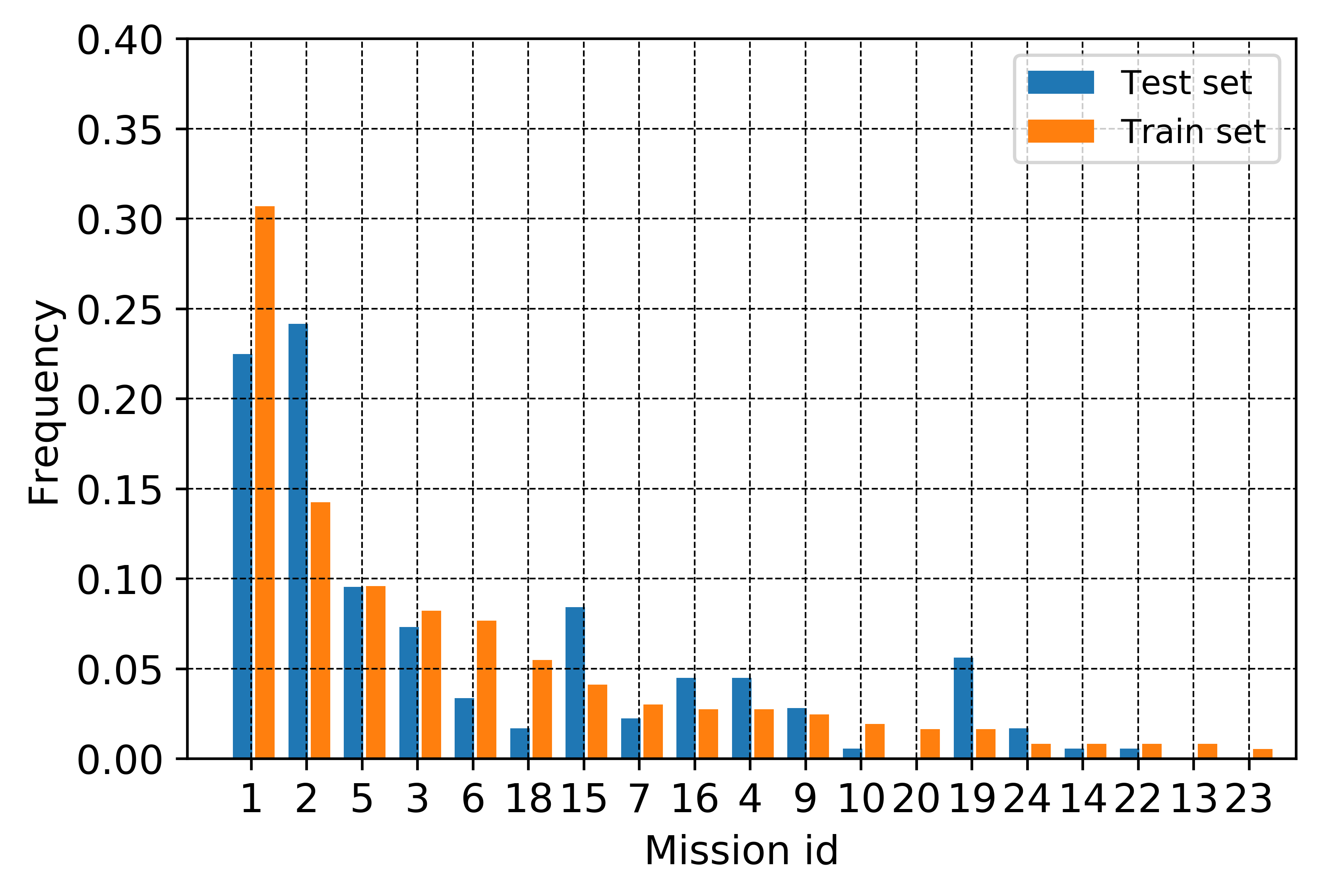}
    \caption{Distribution of the mission type for the test and training sets, for high-risk events.}
    \label{fig:mission_high_risk}
    \end{minipage}
\end{figure}

Therefore, it is important not to create imbalances in mission type when splitting the data into training and test sets. In Figure \ref{fig:mission_all}, we can see that, for the low-risk events, the missions are proportionally represented in both the training and test sets. However, when we only look at the high-risk events, in Figure \ref{fig:mission_high_risk}, we can see that the missions are not well distributed. In particular, \emph{mission\_id: 2}, \emph{mission\_id: 15}, and \emph{mission\_id: 19} are over-represented in the test set, while \emph{mission\_id: 1}, \emph{mission\_id: 6} and \emph{mission\_id: 20} are under-represented. This is because the dataset was randomly split into training and testing, only taking the risk value into account and not the mission type. For future work, it would be recommended to take mission type into account during the splitting of the dataset, as mentioned in Section~\ref{sec:lessons_learned}, or to create datasets with a higher homogeneity with respect to the mission type. Note that further analysis of the dataset split and of the correlation between the training and test sets is conducted in Section \ref{sec:post_comp_ml}.

Another aspect that had to be addressed in the competition design was the choice of the subset of the test set for the purpose of computing the leaderboard (public) score during the competition. We refer to this subset as the \textit{visible} subset. The remaining part of the test set is added at the end of the competition to compute the final ranking and is referred to as the \textit{hold-out} subset. The purpose of only using the \textit{visible} subset to compute the leaderboard score is primarily to prevent teams from overfitting and is a standard setup in machine learning competitions. Furthermore, as a side effect, it adds uncertainty and excitement to the final ranking, incentivizing teams to improve their solutions until the last days. 
For a reasonable submission (e.g. LRP), one would expect a small difference in score when using the \textit{visible} subset or the full test set. In general, if the test set is large enough and the classes are well balanced, one can randomly split the test set into a \textit{visible} and an \textit{hold-out} part. However, as previously mentioned, this was not the case for this dataset. 

Therefore, a simple experiment was conducted to make an informed decision on the \textit{visible} subset selection. Various proportions of low-risk and high-risk events, denoted by $p_{\textrm{low}}$ and $p_{\textrm{high}}$ respectively, were tried. For each of the possible 25 sets of parameters, $p_{\textrm{low}}$, $p_{\textrm{high}}$ $\in \{0.5, 0.6, 0.7, 0.8, 0.9\}$, 200 \textit{visible} subsets were drawn at random and the average score of an LRP forecast computed and compared to $L=0.694$. This simple experiment revealed very clearly how varying $p_{\textrm{high}}$ leads to significant changes in the average score. In particular, for $p_{\textrm{high}}=0.5$, the average change in score on the \textit{visible} subset and the full test set is $10\%$, and drops to $3.5\%$ for $p_{\textrm{high}}=0.9$. The latter variance was considered to be low enough for our purposes and we thus decided to keep $p_{\textrm{high}}=0.9$, and $p_{\textrm{low}}=0.9$ in the \textit{visible} subset. The number of high-risk events that remains in the \textit{hold-out} subset is thus low, introducing the concrete possibility that teams could overfit the \textit{visible} test set. Limiting the number of submission to one per day was considered appropriate to alleviate this risk.

\section{Competition Results}
\label{sec:results}

After the competition ended and the final rankings were made public, a survey was distributed to all the participating teams, in the form of a questionnaire. The result of the survey, the final rankings, the methods of a few of the best ranking teams as well as a brief meta-analysis of all the solutions are reported in this section.

\subsection{Survey}

\begin{wraptable}{R}{0.5\textwidth}
    \centering
    \caption{Background of the participants, out of 22 respondents to the end of the competition questionnaire.}
    \begin{tabular}{c|ccc}
    \toprule
        \multicolumn{1}{c|}{\textbf{Discipline}} & \multicolumn{3}{c}{\textbf{Proficiency}}      \\
          & Professional & Student & Amateur                   \\ \midrule
    Machine Learning & 10 & 10 & 4                                   \\
    Orbital Mechanics & 4  & 5 & 15                                  \\
    \bottomrule
    \end{tabular}
    \label{table:background}
\end{wraptable}

A total of 22 teams participated to the survey, including all the top ranked teams. Some questions from the survey were targeted to gather more information on the background of the participants. 
The questions were phrased as: "How would you describe your knowledge in space debris/astrodynamics?" and a similar one for machine learning data/science. 
The possible answers were limited to "professional", "studying" and "amateur".
The answers are reported In Table \ref{table:background}, where it can be seen that most participants had a background in machine learning, and less in orbital mechanics. Note that the top three teams all identified themselves as machine learning professionals, and two as studying orbital mechanics. 

As mentioned in Section \ref{sec:design} and reported in Table \ref{tab:dataset}, the dataset for the collision avoidance challenge is highly unbalanced, with the train and test sets not randomly sampled from the dataset as mentioned in Sec.~\ref{sec:test_train}. 
A question from the survey probed whether the participants explicitly attempted to deal with the class imbalance (e.g. by artificially balancing the classes, assigning importance weighing to samples, etc) by asking: "Did you apply any approach to compensate for the imbalances in the dataset?".
$65\%$ of the participants answered positively.
Furthermore, half of the participants reported to have attempted to build a validation set with similar properties and risk distribution as the test set, albeit failing since most surveyed teams lamented a poor correlation between training and test sets performances.

One of the main scientific questions that this challenge aimed at tackling is whether the temporal information contained in the time series of CDMs is of use in order to infer the future risk of collision between the target and the chaser. 
A specific question from the survey asked participants if they found the evolution of the attributes over time useful to predict the final risk value.
Perhaps surprisingly, $65\%$ of the teams framed the learning problem as a static one, summarizing the information contained in the time series as an aggregation of attributes (e.g. using summary statistics, or simply the latest available CDM). This may be a direct consequence of the great predictive strength of a naive forecast for this dataset, as outlined in the approaches taken by the top teams, in Sec.~\ref{sec:top_three}. 

Finally, due to the small number of high-risk events in the test set and the emphasis put on false negatives induced by the $F_2$ score, it is natural to ask whether teams probed the test set through a process of trial-and-error. Overall, $30\%$ of the participants (including the top ranked team \textit{sesc}, see Section \ref{sec:top_three}) reported utilising a trial-and-error method to identify high-risk events, suggesting that the difference between the test and training sets posed a serious problem to many teams, a fact that deserved some further insight which we give in Section \ref{sec:post_comp_ml}.

\newpage
 
\subsection{Final rankings}

\begin{figure}[t]
    \centering
    \includegraphics[width=0.55\linewidth]{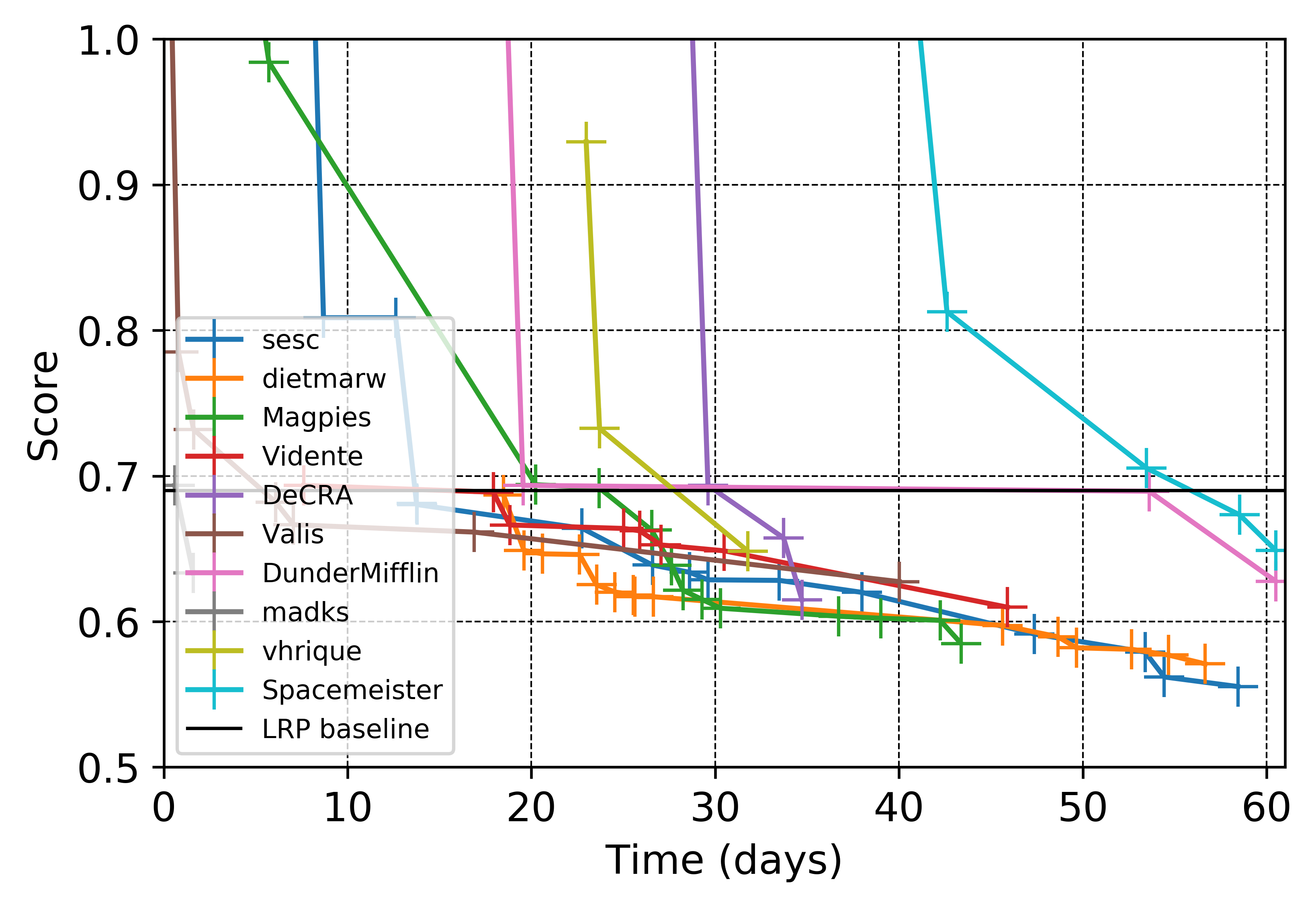}
    \caption{Evolution of the scores of the various top teams submissions.}
    \label{fig:score_evolution}
\end{figure}

\begin{wraptable}{R}{0.45\textwidth}
    \centering
    \caption{Final rankings (from best to worst) evaluated on the test set, for the top 10 teams. The best results are displayed in bold.}
    \label{tab:ranking}
    \begin{tabular}{l|rrr}
        \toprule
        \textbf{Teams}     &     \textbf{Score} & $\textbf{MSE}_{\textbf{HR}}$ & $\mathbf{F_2}$ \\
        \midrule
        LRP baseline           &  0.694   &  0.513   & 0.739  \\
        sesc     &   \textbf{0.556}   &  \textbf{0.407}   & 0.733  \\
        dietmarw    &  0.571 &  0.437   & \textbf{0.765}  \\
        Magpies     &  0.585   &  0.441   & 0.753  \\
        Vidente  &  0.610   &  0.436   & 0.714  \\
        DeCRA &  0.615   &  0.457   & 0.743  \\
        Valis   &  0.628   &  0.467   & 0.744  \\
        DunderMifflin     &  0.628   &  0.451   & 0.718  \\
        madks     &  0.634   &  0.476   & 0.750  \\
        vhrique     &  0.649   &  0.496   & 0.764  \\
        Spacemeister     &  0.649   &  0.479   & 0.738  \\
        \bottomrule
    \end{tabular}
\end{wraptable}

96 teams participated to the challenge and produced a total of 862 different submissions during the competition timeframe.
The scores on the leaderboard changed frequently and the final ranking remained uncertain up until the end of the competition. The evolution of the scores for the top 10 teams over the course of the competition is reported in Figure \ref{fig:score_evolution}. It is interesting to note how the top four teams closely competed for first place up to the very last days.
Another observation is that while all the top teams managed to beat the LRP baseline, it took most of the teams around 20 days to do so, implying that the LRP baseline was fairly strong.
This is further supported by the fact that the scores do not improve much below the LRP baseline, hinting that the naive forecast is an important predictor of the risk value at closest approach.

The final results, broken down into the $\textrm{MSE}_{\textrm{HR}}$, with the risk clipped at -6.001 and the $F_2$ components are shown in Table \ref{tab:ranking}, for the top 10 teams.
All the teams managed to improve upon the LRP baseline score by obtaining a better $\textrm{MSE}_{\textrm{HR}}$. Interestingly, however, many of the teams fail to obtain a better $F_2$ value than the LRP baseline. 

\begin{figure*}[tb]
    \subfloat[]{
    	\includegraphics[width=0.5\linewidth]{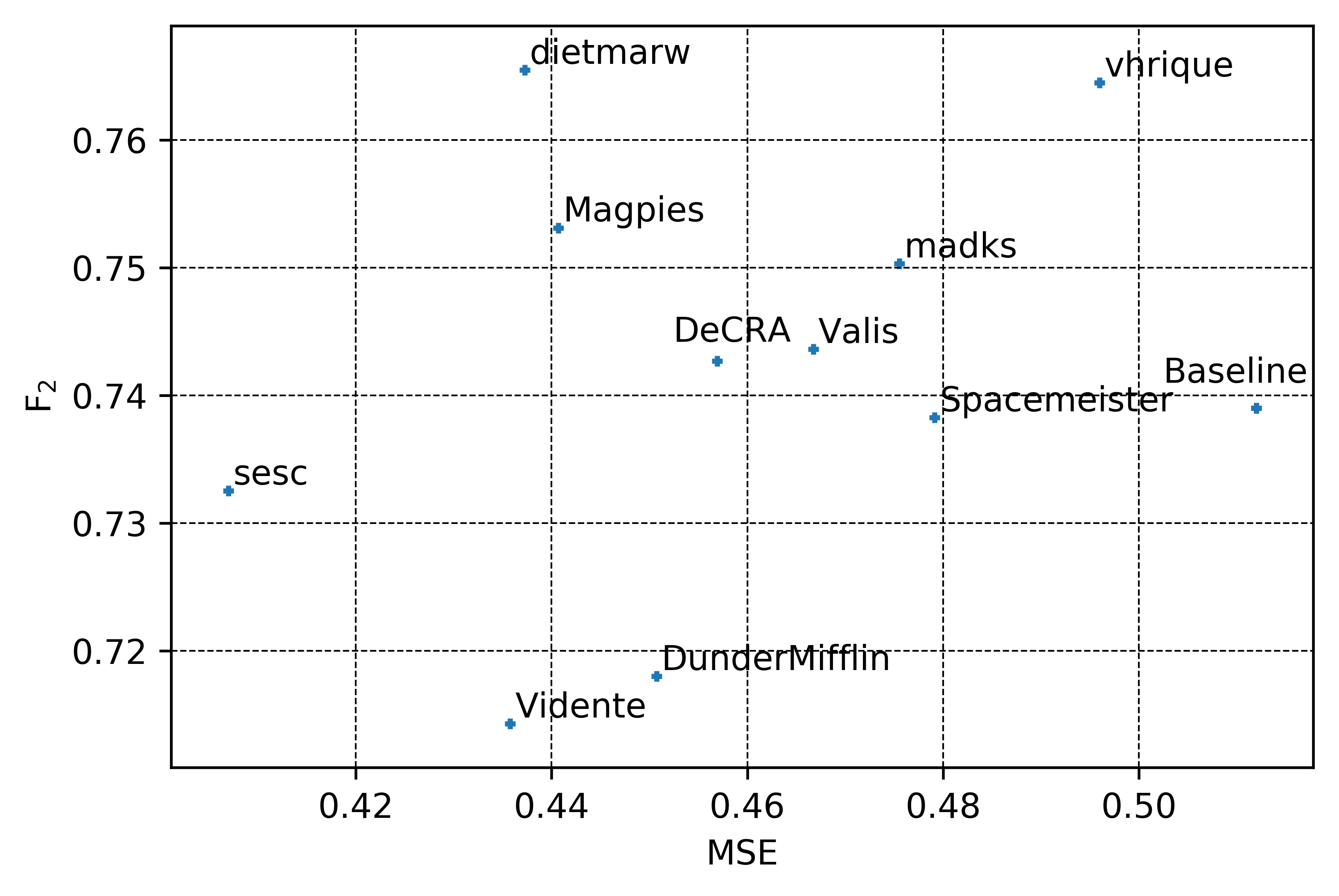}}
    \subfloat[]{
    	\includegraphics[width=0.5\linewidth]{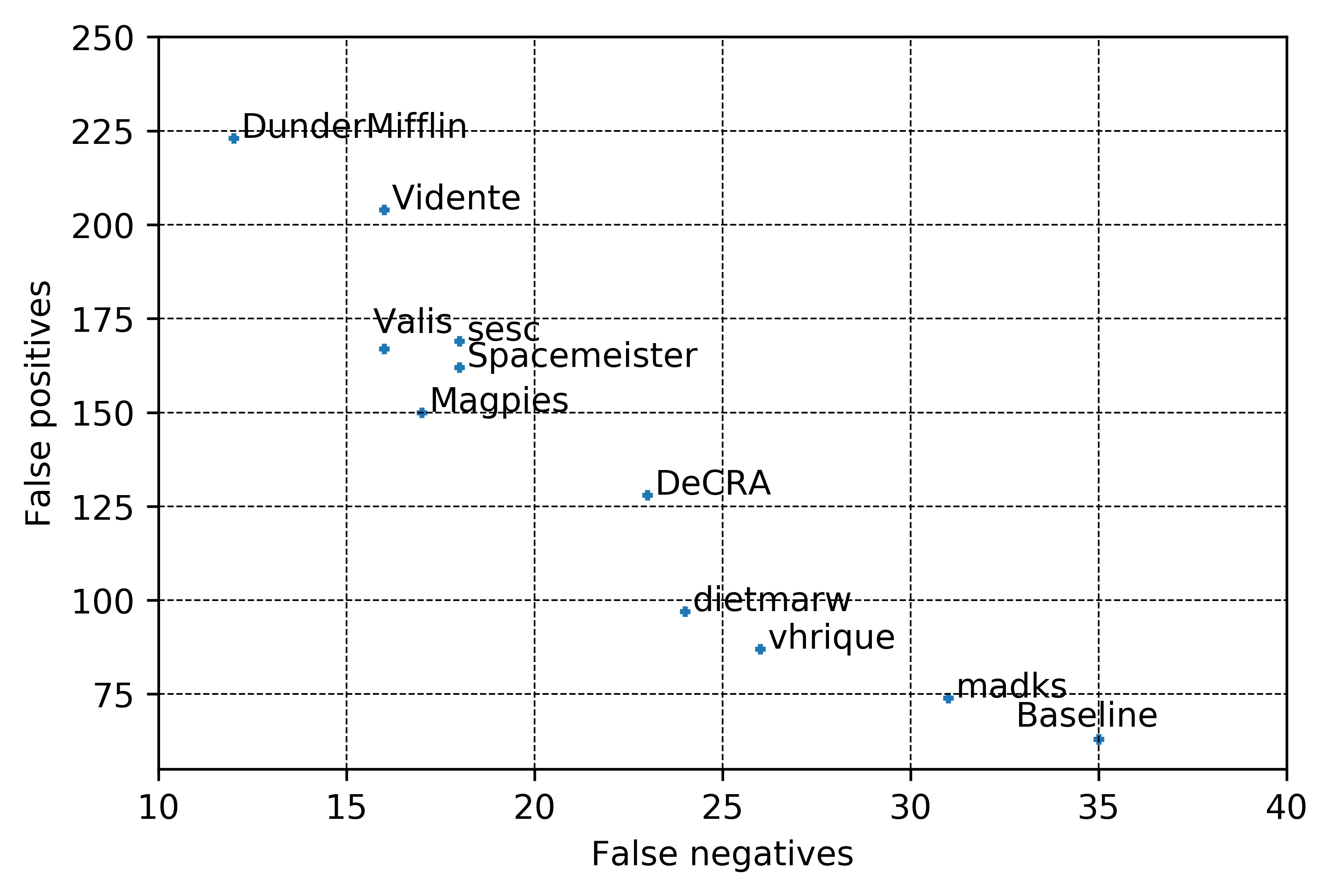}}
    \caption{On the left, in (a), the $F_2$ score and the $\textrm{MSE}_{\textrm{HR}}$ are plotted for the top 10 teams. On the right, in (b), the $F_2$ scores is broken into two components: false negatives (out of 150 positive events) and false positives (out of 2017 negative events).}
    \label{fig:pareto_scores}
\end{figure*}

In order to investigate further the differences between the $F_2$ score achieved by the teams and the LRP baseline solution, it is helpful to look into the false positives and false negatives of each returned model, as reported in Figure \ref{fig:pareto_scores} (b). We can see that the Pareto front is very heterogeneous and consists of several teams: \emph{DunderMifflin}, \emph{Valis}, \emph{Magpies}, \emph{DeCRA}, \emph{dietmarw}, \emph{vhrique}, \emph{madks} and the baseline solution, denoted as \emph{Baseline}. Even though the baseline solution is in the Pareto front, we can see that the resulting $F_2$ score, in (a), is dominated by several teams. This is due to the fact that the $F_2$ score puts more emphasis on penalizing false negatives which the baseline solution has the most of. In (a), only two teams remain in the Pareto front: \emph{sesc}, and \emph{dietmarw}. Interestingly, \emph{dietmarw} has the highest $F_2$ score and \emph{sesc} has the lowest $\textrm{MSE}_{\textrm{HR}}$, suggesting that their methods could be combined in order to achieve a better overall score.

\subsection{Methods used by the $1^{\textrm{st}}$ and $3^{\textrm{rd}}$ ranked teams}
\label{sec:top_three}

\subsubsection{Team sesc}
\label{sec:sesc}



\begin{table}[t]
    \fontsize{9.5}{9.5}\selectfont
     \centering
     \caption{Evaluation of team sesc's approach, as additional steps were added and combinations tested.
     Loss evaluations on the competition's public leaderboard included.
     Risk values clipped at -6.00001 (-6.001 in LRP).}
     \label{tab:manual_eng}
\begin{tabular}{lrrr|rrrr}
\toprule
{\textbf{Combinations of steps}} & \multicolumn{3}{c|}{\textbf{Train set}} & \multicolumn{4}{c}{\textbf{Test set}} \\
{} &       \MSE &     $F_2$ &   loss &      \MSE &     $F_2$ &   loss & leaderboard \\
\midrule
LRP baseline                                     &     0.330 &  0.411 &  0.804 &    0.513 &  0.739 &  0.694 &       0.718 \\
steps: 0 (raise to -5.95)                        &     0.330 &  0.430 &  0.768 &    0.512 &  0.753 &  0.680 &       0.703 \\
steps: 0 + 1 (raise to -5.60)                    &     0.305 &  0.392 &  0.779 &    0.498 &  0.764 &  0.653 &       0.670 \\
steps: 0 + 1 + 2 (raise to -5.00)                &     0.290 &  0.296 &  0.982 &    0.445 &  0.738 &  0.603 &       0.612 \\
steps: 0 + 1 + 2 + 3 (safe c\_object\_type)        &     0.290 &  0.301 &  0.966 &    0.426 &  0.735 &  0.579 &       0.587 \\
steps: 0 + 1 + 2 + 4 (safe t\_span)               &     0.290 &  0.298 &  0.974 &    0.447 &  0.735 &  0.608 &       0.611 \\
steps: 0 + 1 + 2 + 5 (safe miss\_distance)        &     0.325 &  0.304 &  1.070 &    0.444 &  0.733 &  0.607 &       0.613 \\
steps: 0 + 1 + 2 + 6 (clip high risks)           &     0.293 &  0.296 &  0.990 &    0.424 &  0.738 &  0.575 &       0.581 \\
steps: 0 + 1 + 2 + 3 + 4 + 5 + 6                 &     0.327 &  0.311 &  1.050 &    0.414 &  0.728 &  0.569 &       0.564 \\
steps: 0 + 1 + 2 + 5 + 6                         &     0.327 &  0.304 &  1.077 &    0.424 &  0.733 &  0.578 &       0.581 \\
steps: 0 + 1 + 2 + 5 + 6 + 7 (manual adjustment) &     0.327 &  0.304 &  1.077 &    0.407 &  0.733 &  0.555 &       0.555 \\
\bottomrule
\end{tabular}
\end{table}

The highest ranking team was composed of scientists coming from diverse domains of expertise: evolutionary optimization, machine learning, computer vision, data science and energy management. In the early stages of the competition, the team attempted to use different methods including extracting time series features \cite{tsfresh_url}, constructing an automated machine learning pipeline via Genetic Programming \cite{OlsonGECCO2016,Wang2019}), and using random forests. 
All these approaches were reported to have a score of $L \in[0.83, 1.0]$ on the test set, but performed radically better on the training set. Such a difference was taken as an indication that an automated, off-the-shelf machine learning pipeline was likely not to be the appropriate way to learn from this dataset.

Instead, the team resorted to a step-by-step approach informed by statistical analysis and taking advantage of the metric and the constitution of the test set. Indeed, the $F_2$ score is biased towards false negatives and there is a relatively higher proportion of high-risk events in the test set than in the train set. Furthermore, it can be observed that, in the training set, most of the high-risk events misclassified by the naive forecast have a latest risk $r_{-2}$ only slightly below the threshold. A then simple strategy is to promote borderline low-risk events to high-risk events, thus improving recall (at the cost of penalizing precision), which is what the $F_2$ score puts emphasis on. In practice, this strategy was implemented by introducing three thresholds, referred to as \textit{step 0}, \textit{step 1} and \textit{step 2}, as shown in Table \ref{tab:manual_eng} and Eq.~(\ref{eq:thresholds}).

Additional incremental improvements were achieved by assigning events to low risk whenever either the chaser type (\textit{c\_object\_type} attribute) was identified as a payload, the diameter of the satellite (\textit{t\_span} attribute) was small (below $0.5$) or the \textit{miss\_distance} was above $30000$ meters. These steps are referred to as \textit{step 3}, \textit{step 4} and \textit{step 5} respectively, in Table \ref{tab:manual_eng} and Eq.~(\ref{eq:thresholds}).

Finally, the risk value for high-risk events was clipped to a slightly lower risk value to enforce the general trend of risk decrease over time, thus improving the $\textrm{MSE}_{HR}$ while preserving the $F_2$ score. This step is referred to as \textit{step 6} in Table \ref{tab:manual_eng} and Eq.~(\ref{eq:thresholds}).
    
In summary, the aforementioned observations led to the introduction of a cascade of thresholds:

\begin{align}
\label{eq:thresholds}
   \hat r =
        \begin{cases}
            -5.95    \phantom{0001} \text{if } -6.04 \leq r_{-2} < -6.00                & (\textit{step 0}), \\
            -5.60    \phantom{0001} \text{if } -6.40 \leq r_{-2} < -6.04                & (\textit{step 1}), \\
            -5.00    \phantom{0001} \text{if } -7.30 \leq r_{-2} < -6.40                & (\textit{step 2}), \\
            -6.00001 \phantom{0}    \text{if } c\_object\_type \text{ is ``payload''}   & (\textit{step 3}), \\
            -6.00001 \phantom{0}    \text{if } t\_span < 0.5                            & (\textit{step 4}), \\
            -6.00001 \phantom{0}    \text{if } miss\_distance > 30000                   & (\textit{step 5}), \\
            -4.00    \phantom{0001} \text{if } -4.00 \leq r_{-2} < -3.50                & (\textit{step 6}), \\
            -3.50    \phantom{0001} \text{if } \phantom{-4.00 \leq r} r_{-2} \geq -3.50 & (\textit{step 6}).
        \end{cases}
\end{align}

\subsubsection{Team Magpies}

The third ranked team was made of a single Space Situational Awareness (SSA) researcher and Machine Learning (ML) engineer. The team achieved its final score by leveraging Manhattan-LSTMs~\cite{muller2016siamese} (Figure~\ref{fig:man_lstm}), a siamese architecture based on recurrent neural networks. Team Magpies started with analysing the dataset and filtered the training data according to the test set requirements laid out in Section \ref{sec:test_train}. A comprehensive exploratory data analysis was conducted and provided the following conclusions:

\begin{figure}[tbh]
    \centering
    \includegraphics[width=0.75\linewidth]{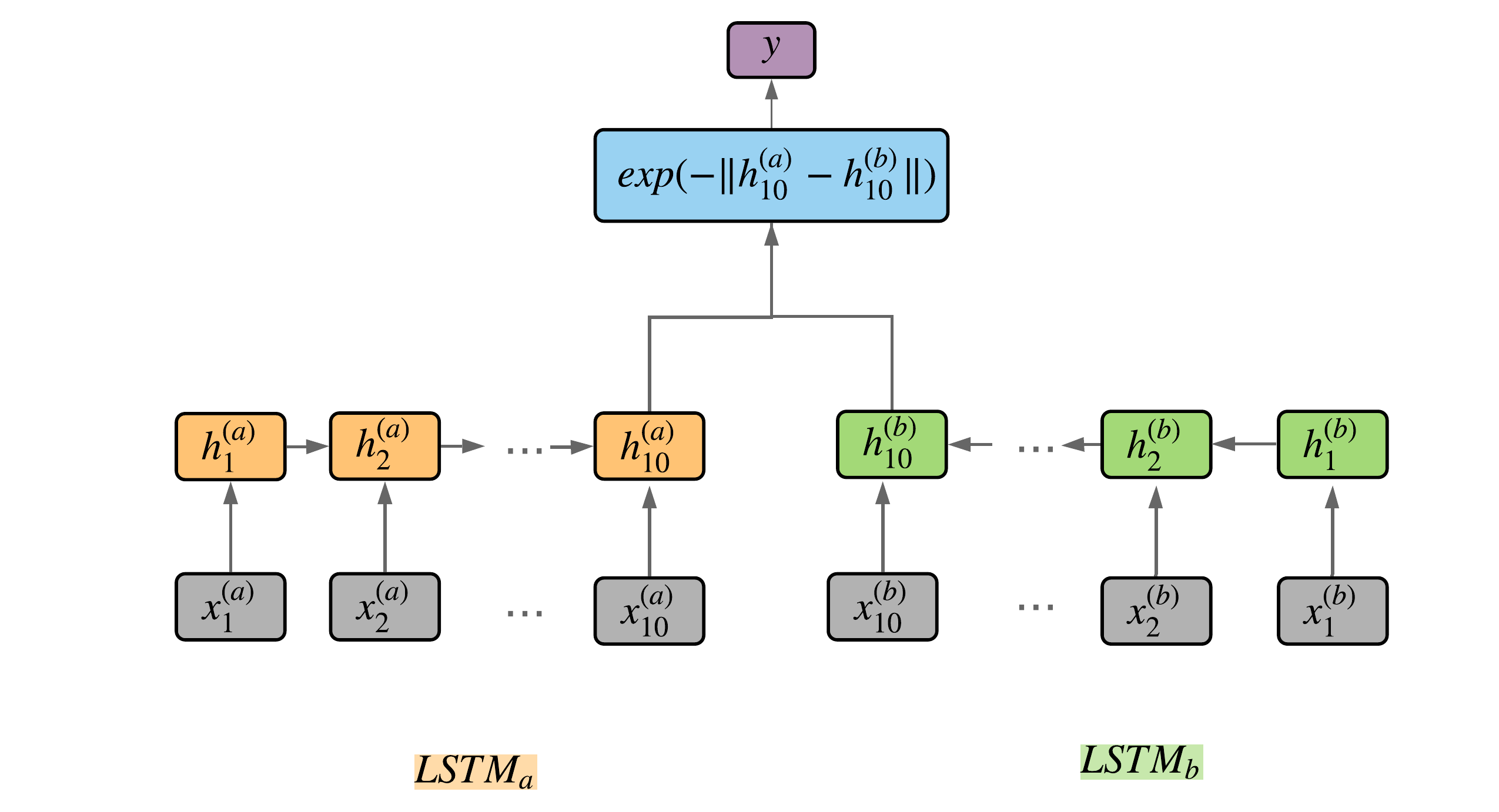}
    \caption{Diagram showing the Manhattan-LSTM architecture.}
    \label{fig:man_lstm}
\end{figure}

\begin{wrapfigure}{R}{0.5\textwidth}
    \centering
    \includegraphics[width=0.98\linewidth]{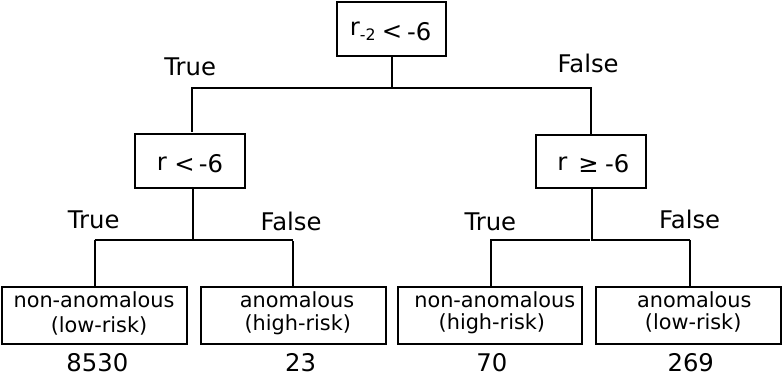}
    \caption{Diagram showing the number of anomalous and non-anomalous events from the train set, for the two cases: (i) left root split is low-to-high risk anomalies; (ii) right root split is high-to-low anomalies.
    Event counts (shown underneath the tree leaves) are based on filtering events as per test set specifications in Section~\ref{sec:test_train}, but allowing for final CDMs to lie within 2 days of the TCA.}
    \label{fig:anomalyfig}
\end{wrapfigure}

\begin{itemize}
    \item It is clear that there exists a distribution difference in the training and test sets, as highlighted in Section \ref{sec:post_comp_ml} (in particular with the proportion of low-risk and high-risk events), making the latest available risk value (LRP baseline) is a strong solution.
     \item The dataset is highly imbalanced, which makes it hard to learn reliable signal from the small number of high-risk events in the training set. Similarity-based models are promising as they can help alleviate the class imbalance problem \cite{rasit2020autocollision, buitrago2018anomaly}. One particular approach is to use siamese networks as they have been shown to be an efficient way to learn similarities between inputs coming from previously unseen classes, in few-shots and one-shot learning settings \cite{koch2015siamese}.
     \item Picking the features that have strong predictive power is important not to learn noise in the data and overfit on the training set. Therefore, 7 out of 103 features (\emph{time\_to\_tca}, \emph{max\_risk\_estimate}, \emph{max\_risk\_scaling}, \emph{mahalanobis\_distance}, \emph{miss\_distance}, \emph{c\_position\_covariance\_det}, and \emph{c\_obs\_used}) are selected by comparing the distribution difference of the non-anomalous event (last available collision risk is low and ends up low at close approach, and vice-versa for high-risk events) and anomalous ones (last available collision risk is low and ends up high at close approach, and vice-versa for high-risk events). For these 7 feature, the attributes come from the latest available CDM. The number of arising anomalous and non-anomalous cases are shown in Figure \ref{fig:anomalyfig}. Note that we do not attempt to identify the anomalous low-risk events (i.e. events that go from high risk to low risk). This is due to the fact that these events are harder to discriminate. In addition, the metric does not strongly penalize false positives. 
     \item In addition to those 7 attributes, three new features are added: the number of CDMs (\emph{number\_CDMs}) issued before 2 days, the mean (\emph{mean\_risk\_CDMs}) and standard deviation (\emph{std\_risk\_CDMs}) of the risk values of those CDMs. These features allow to capture information from the CDMs time series without encoding any temporal information (e.g. rate of change, moving average, etc).
\end{itemize}

Next, we introduce the training and evaluation protocol of the Manhattan-LSTM in order to exploit the distribution difference.
The activation functions are hyperbolic tangents and Adam \cite{kingma2014adam} is used as the gradient descent optimizer. The training data is split using a 3-fold cross-validation (8 events are selected in each validation fold, from the 23 anomalous events). Then, $\{\textrm{non-anomalous}, \textrm{non-anomalous}\}$ and $\{\textrm{non-anomalous}, \textrm{anomalous}\}$ pairs are generated for the siamese network to learn similar and dissimilar pairs, respectively. During the training phase, anomalous events are over-sampled in order to balance the number of anomalous and normal pairs. For each validation fold, several networks are trained with different hyperparameter regimes, uniformly sampled from the discrete set $\{\textit{minimum}, \textit{minimum}+\textit{step}, \ldots, \textit{maximum}\}$, shown in Table~\ref{tab:hyperparameters}. Networks that attain a reasonable enough performance (area under the curve (AUC) value of at least $0.8$) when evaluated on the 3 validation folds are then selected in a majority voting ensemble scheme, with equal weight. \\

To summarise, let us denote the output of the majority vote by $f$, which takes as input 10 features, denoted by $x$ and predicts whether a low-risk event is anomalous or not. Then, the final predictions on the test set are given by

\begin{equation}
\label{eq:rasit_model}
   \hat r =
        \begin{cases}
            -6.001 \hspace{0.1cm} \text{if} \hspace{0.1cm} r_{-2} < -6 \hspace{0.1cm} \text{and} \hspace{0.1cm} f(x) = \text{non-anomalous}, \\
            -5.35 \hspace{0.29cm} \text{if} \hspace{0.1cm} r_{-2} < -6 \hspace{0.1cm} \text{and} \hspace{0.1cm} f(x) = \text{anomalous}, \\
            \hspace{0.3cm} r_{-2} \hspace{0.35cm} \text{if} \hspace{0.1cm} r_{-2} \geq -6,
        \end{cases}
\end{equation}
where $-5.35$ is the average risk value of all the high-risk events in the training set. Overall, this approach achieves a $F_2$ score of 0.753 and a $MSE_{HR}$ loss of 0.441, on the test set (see Table \ref{tab:ranking}). 

\begin{table}[b]
    \centering
    \caption{Discrete search space for sampling the hyperparameter values of the Manhattan-LSTM models.}
    \label{tab:hyperparameters}
    \begin{tabular}{l|ccc}
        \toprule
        \textbf{Hyperparameters} & \textbf{Minimum} & \textbf{Maximum} & \textbf{Step size} \\
        \midrule
        Number of hidden units    &  8   &  128  & 8 \\
        Gradient clipping   &  0.0   &  0.5   & 0.01 \\
        Batch size   &  16   &  512   & 16\\
        Learning rate   &  $2 \cdot 10^{-6}$   & $10^{-4}$ & $10^{-6}$ \\                
        \bottomrule
    \end{tabular}
\end{table}


\subsection{Difficulty of samples}

\begin{figure*}[tb]
    \subfloat[]{
    	\includegraphics[width=0.48\linewidth]{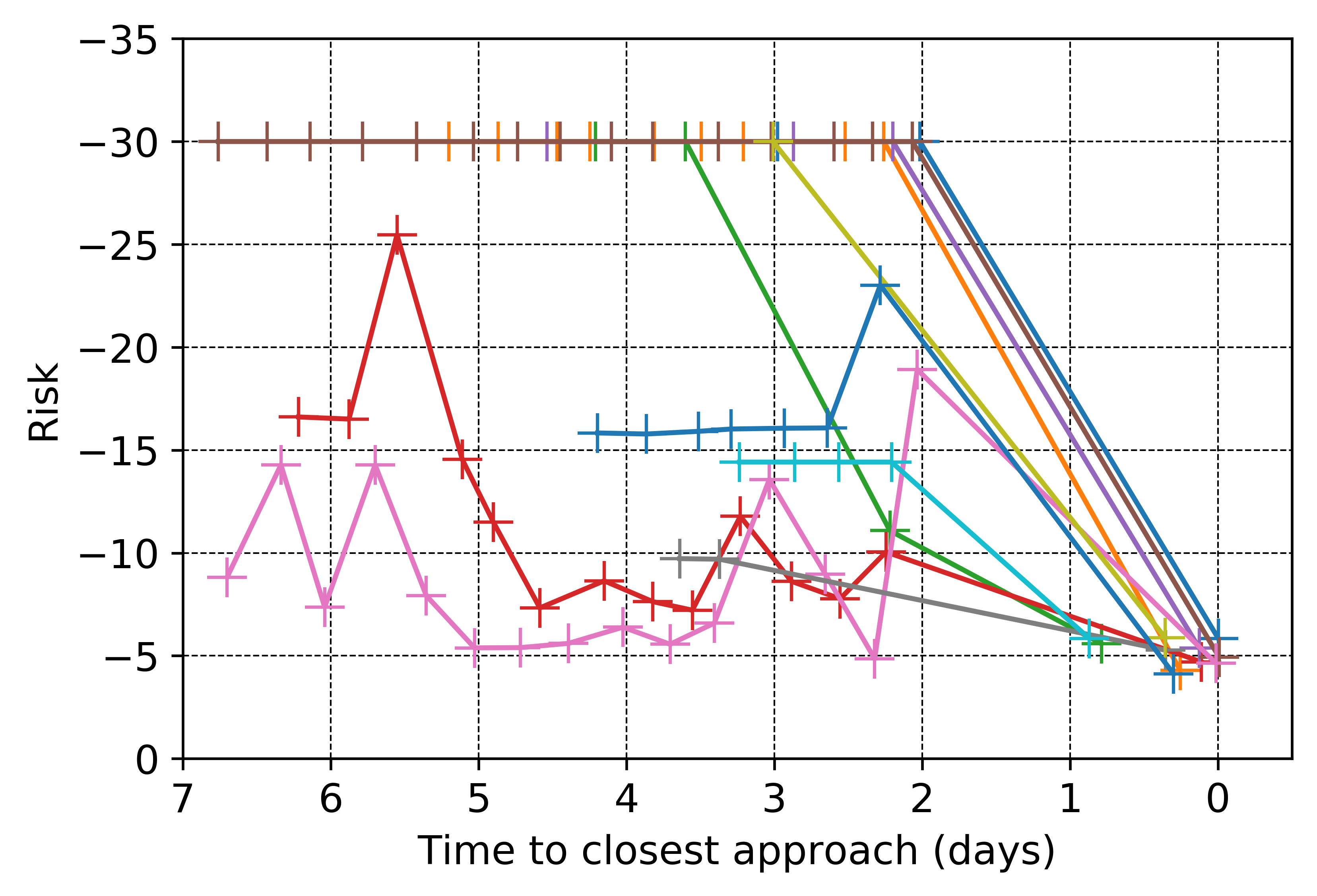}}
    \subfloat[]{
    	\includegraphics[width=0.48\linewidth]{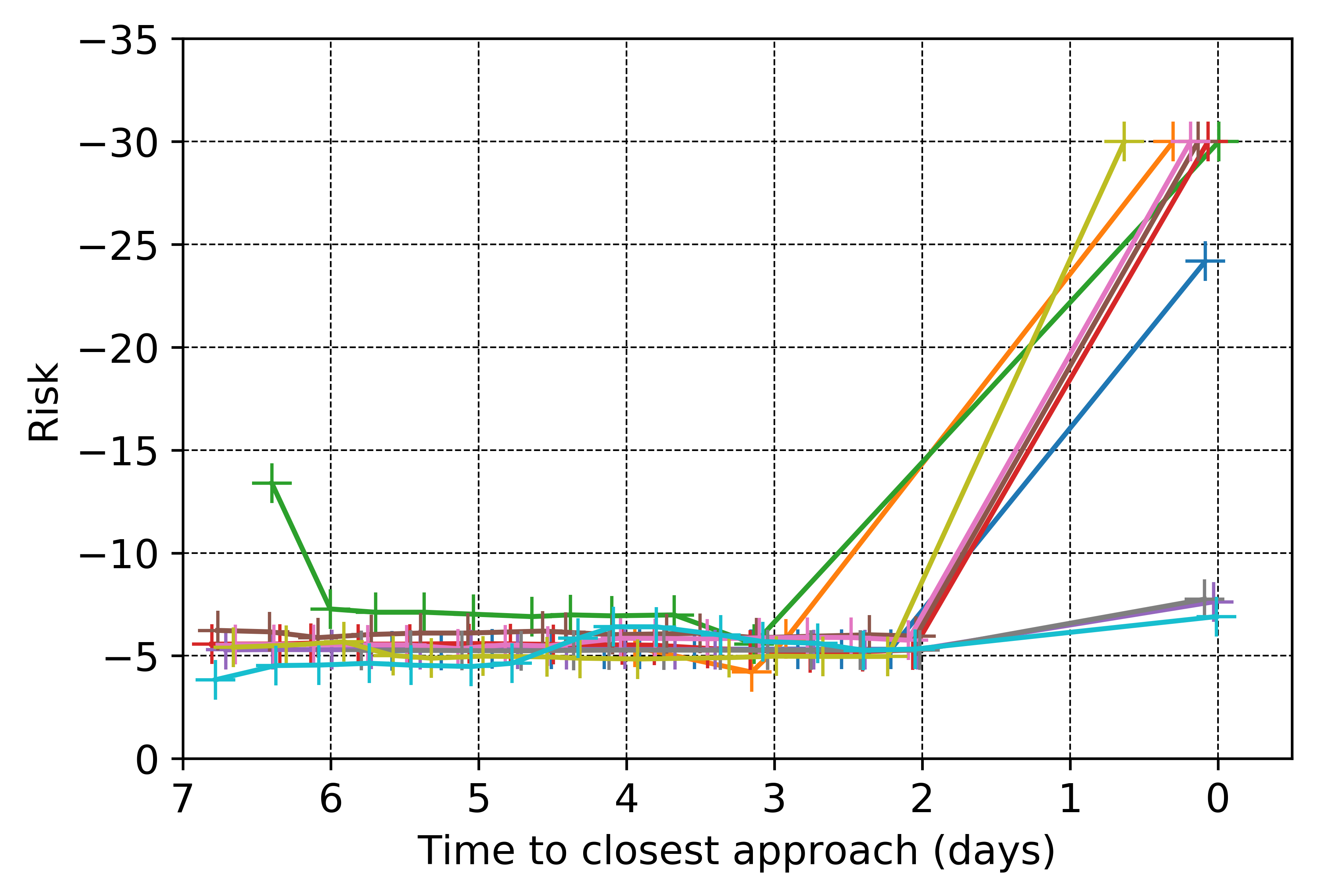}}
    \caption{Events consistently misclassified by all the top 10 teams: (a) false negatives, (b) false positives. On the left panel, we show all the false negatives (11 out of 150 high risk events). Each event is represented as a line and the CDMs are marked with crosses. The evolution of risk between two CDMs is simply plotted as a linear interpolation. On the right panel, we show 10 randomly sampled events, out of 62 false positives in total. These events are particularly hard to classify due to the big jump in risk as we are getting closer to time of closest approach, going from low risk to high risk in (a) and vice-versa in (b).}
    \label{fig:mispred}
\end{figure*}

\begin{wrapfigure}[18]{R}{0.5\textwidth}
    \includegraphics[width=0.98\linewidth]{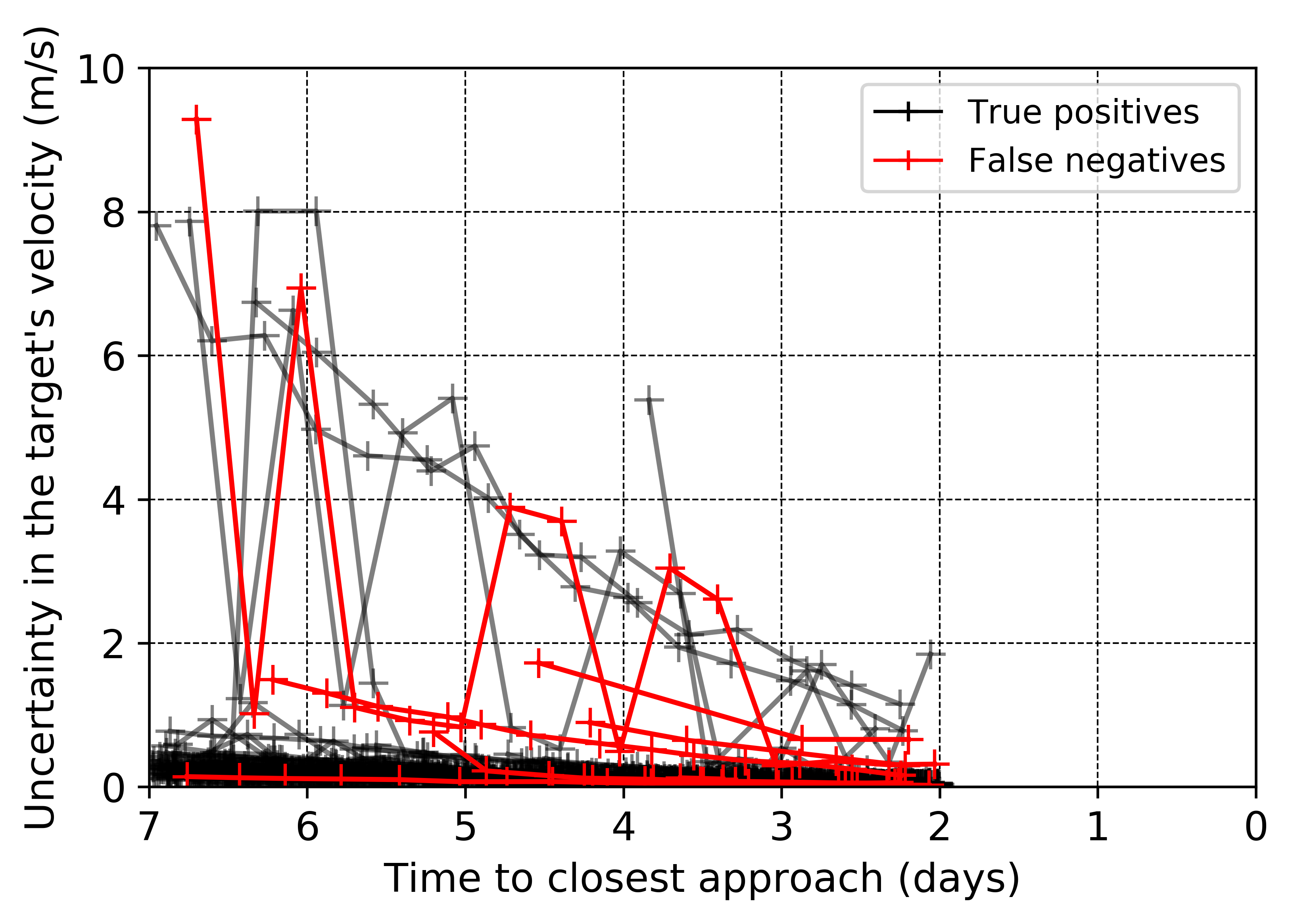}
    \caption{Evolution of the uncertainty in the radial velocity of the target spacecraft (\emph{t\_sigma\_rdot}) over time, up until 2 days to time of closest approach. The evolution of the uncertainty of the 11 false negative events (see Figure \ref{fig:mispred}) is shown in red. The evolution of the uncertainty of the 139 remaining true positive events is shown in black.}
    \label{fig:cov_high_risk}
\end{wrapfigure}

In this section, we investigate the events in the test set that were consistently misclassified by all the top 10 teams. These events can be separated into two groups: false positives and false negatives. False negatives correspond to events that were wrongly classified as low risk and false positives to events wrongly classified as high risk. Figure~\ref{fig:mispred} shows the evolution of risk for events that were consistently misclassified. We can see that those events all experience a significant change in their risk value, as they progress to closest approach, thus rendering the use of the latest risk value misleading. Furthermore, as seen from Figure~\ref{fig:mispred}, the temporal information is likely to be of little use to make good inferences in these cases: there is no visible trend and the risk value jumps from one extreme to the other (from very low risk to very high risk in (a) and vice-versa in (b)). One characteristic that all these events have in common is high uncertainties in their associated measurements (e.g. position, velocity), leading to very uncertain risk estimates, susceptible to big jumps close to TCA. In Figure \ref{fig:cov_high_risk}, the evolution of the uncertainty in the radial velocity of the target spacecraft (\emph{t\_sigma\_rdot}) is displayed, for the the 150 high-risk events in the test set. 

It can be seen that the uncertainty values are generally higher for the misclassified events. Note that there are many more uncertainty attributes recorded in the dataset and Figure \ref{fig:cov_high_risk} is only showing one of them. 
Higher uncertainties for the misclassified events suggests that there there may be value in building a model which takes these uncertainties into account at inference time, by outputting a risk distribution instead of a point-prediction, for instance.

\section{Post competition ML}
\label{sec:post_comp_ml}

Further machine learning experiments were carried out on the dataset, both to analyse the competition and to further investigate the use of predictive models for collision avoidance. The aim of these experiments was to shed light on the difficulties experienced by competitors in this challenge, and to gain deeper insights into the ability of machine learning models to learn generalizable knowledge from this data.

\subsection{Train/test set loss correlation}
\label{sec:gen:official_split}

A first experiment was designed to analyse the correlation between the performances of ML models on the train and test sets used during the competition. Only the train set events conforming to the test set specifications (Section~\ref{sec:test_train}) were considered: final CDM within 1 day of TCA, and all other CDMs at least 2 days away from TCA. The last CDM at least 2 days away from TCA is the most recent CDM available to operators when taking the final planning decisions and was used as the sole input to the model. In other words, temporal information from the CDMs time series is not taken into account. From it, only the numerical features were used: the two categorical features (\textit{mission\_id} and \textit{object\_type}) were discarded. This way, for models to learn mission or object specific rules, they would have to do so through features encoding relevant properties of that feature or object, in the hope this would force the model to learn more generalizable rules. Other than this step, no other transformations were applied over the CDM's raw values (such as scalings or embeddings). Likewise, no steps were taken to impute the missing values that at times occur in many of the CDM's variables. We left these to be handled by the machine learning algorithm (LightGBM in our case), through its own internal mechanisms. Most importantly, the model's target was defined to be the change in risk value between the input CDM and the event's final CDM ($r - r_{-2}$), rather than the final risk $r$ itself. This facilitates the learning task as it implicitly reduces the bias towards the most represented final risk (i.e. $-30$). Furthermore, it allows for a direct comparison to the LRP baseline as the various models are, de facto, tasked to predict a new estimator $h$ such that $r = LRP + h$.
This quantity $h$ was further encoded through a quantile transformer, in order to assume a uniform distribution.

%
%
The training data, eventually, consisted of a table of 8293 CDMs, from as many events, each described by 100 features. Each CDM had assigned to itself a single numerical value that was to be predicted. Overall, these steps result in a rather simplified data pipeline. Note the absence of any steps to address the existing class imbalance and, in addition, the absence of any focus on the high-risk events during the training process. Models are asked to learn the evolution of risk across the full range of risk values, even though, at evaluation time they are mostly assessed on their performance at one end of the range of risk values. Indeed, the competition’s \MSE\ metric is taken only over the true high-risk events, and the use of clipping at a risk of -6.001 further ignores where the final predicted risk lies, if it falls below this value. Also $F_2$ is a classification metric and cares only for where risk values lie with respect to the -6 threshold.

For the kind of regression problem outlined above, with a tabular data representation, gradient boosting methods \cite{elements} offer state-of-the-art performance. We thus chose the LightGBM gradient boosting framework \cite{LightGBM} to train a large number of models.
%
%
To attain both training speed and model diversity, we changed hyperparameters as follows (with respect to the defaults in the \texttt{LGBMRegressor} of LightGBM 2.2.3): the \texttt{n\_estimators} was set to 25, \texttt{feature\_fraction} to 0.25, and \texttt{learning\_rate} to 0.05.
Together, these settings result in an ensemble with fewer decision trees (default is 100), each tree is trained exclusively on a small random subset of 25\% of the available features (default is 100\%), and each successive tree has a reduced capability to overrule what previous trees have learned (default learning rate, also known as shrinkage in the literature, is 0.1).

\begin{figure*}[t]
    \centering
    \includegraphics[width=\textwidth]{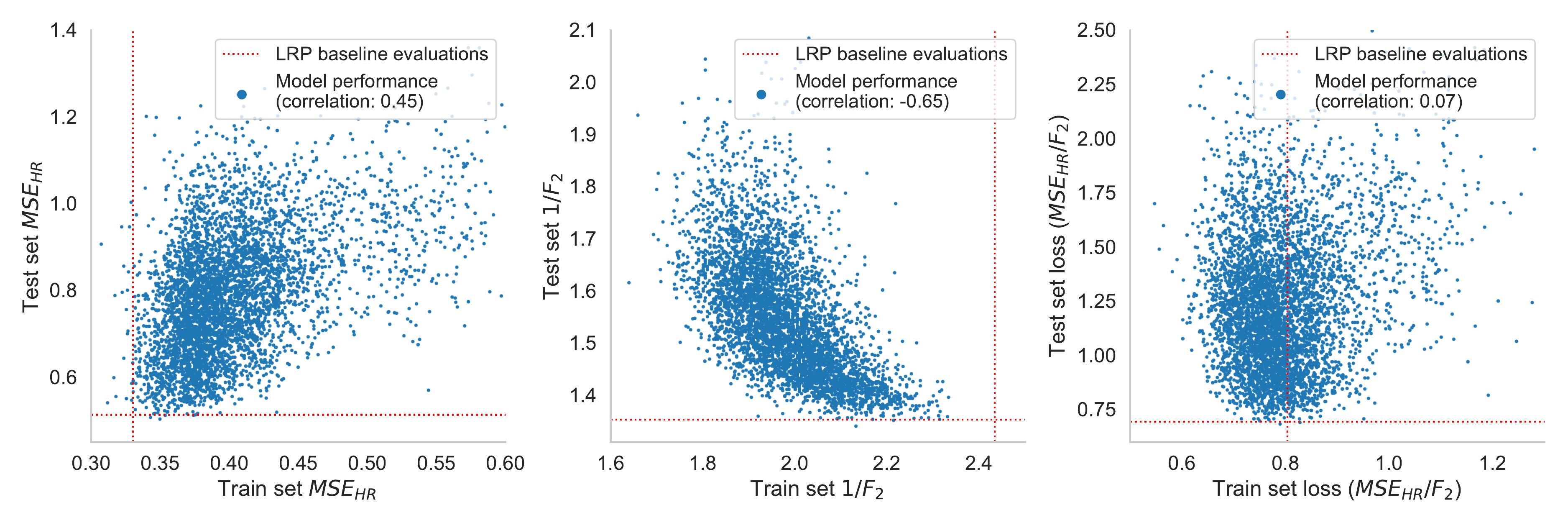}
    \caption{Performance levels achieved by 5000 gradient boosting regression models trained using the competition's dataset split.}
    \label{fig:comp_sim__official_split}
\end{figure*}

We can see in Figure~\ref{fig:comp_sim__official_split} the evaluations of 5000 models, on the train and set sets, on the \MSE\ and $1/F_2$ metrics, as well as their product (the competition’s score, or loss metric). Risk values were clipped at -6.001 prior to measuring \MSE. We compare models’ performance on the train ($x$-axis) and test sets ($y$-axis). As a reference, dotted lines show performance of the LRP baseline (Section~\ref{sec:baseline}). The Spearman rank-order correlation coefficient is computed, as indicator of the extent to which performance in the train set generalizes to the test set. 

Only one model (0.02\% of the trained models) managed to outperform the LRP baseline loss in both the train and test sets. With a test set loss of 0.684 (1.4\% gain over LRP), that model would have ranked 11th in the official competition.
%
%

Overall, Figure~\ref{fig:comp_sim__official_split} displays several undesirable trends.
The \MSE\ plot shows a positive correlation: train set performance is predictive of performance on the test set. However, models struggle to improve on the LRP, in both sets. Most models degrade the risk estimate available in the most recent CDM. In the case of $1/F_2$, we actually see a strong negative correlation: the better performance was on the train set, the worse it was on the test set. This is a clear sign of overfitting. When aggregated, we were thus left with a loss metric displaying essentially no correlation. This observation, while bound to the modelling choices made, offers a possible explanation to competitors' sentiment of disappointment for models that were good in their local training setups evaluating poorly on the leaderboard.

\subsection{Simulating 10,000 virtual competitions}
\label{sec:gen:different_splits}

To further our understanding on the absence of a significant Spearman rank correlation between training and test set performances, as highlighted in Figure \ref{fig:comp_sim__official_split}, we simulated 10000 possible competitions differing in the data split. In each, a test size was randomly chosen from a set of 19 options, containing the values from 0.05 through 0.95 in steps of 0.05. This setting indicates the fraction of events that should be randomly selected to be moved to the test set. The full dataset being partitioned was composed solely of the 10460 events that conform to the official competition’s test set specifications. We adopted here a different splitting procedure from the one reported in Section~\ref{sec:data_split}. A stratified shuffle splitter was used, so the proportions of final high-risk events in both the train and test sets would always match the proportion seen in the dataset being partitioned (2.07\%) as closely as possible.
%
%
For reference, a test size of 0.2 leaves on average 172.8 high-risk events in the train set, and 43.2 in the test set (and 8195.2 and 2048.8 low-risk events in each, respectively).
%
No allowances were made to preserve in the train and test sets the events’ distributions of mission ids and chaser object types present in the full dataset being partitioned. As shown in Figures~\ref{fig:mission_all}--\ref{fig:mission_high_risk}, the fraction of events from the different missions has such an imbalance that many of these generated splits likely either leave some missions wholly unrepresented in either the train or test set, or with such low volumes as to render the learning of their properties unlikely. Likewise, the object types attribute has an identical imbalance and will therefore pose similar challenges. Although this choice makes it more difficult to achieve a higher score on the performance metrics, it serves the current purpose of evaluating generalization capability.

In each of the 10000 virtual competitions, 100 regression models were trained, using the same data pipeline and model settings as described in the previous section’s experiment. On average, 526 competitions were simulated per each of the 19 different test size settings, each with its own distinct data split. In total, 1 million models were trained.
Although framed here as virtual recreations of the Kelvins competition, this process actually implements, per test size setting, a Monte Carlo cross-validation, or repeated random sub-sampling validation \cite{Kuhn2013}.
Were the number of random data splits to approach infinity, results would thus tend towards those of leave-p-out cross-validation.
%

\begin{figure*}[t]
    \centering
    \includegraphics[width=\textwidth]{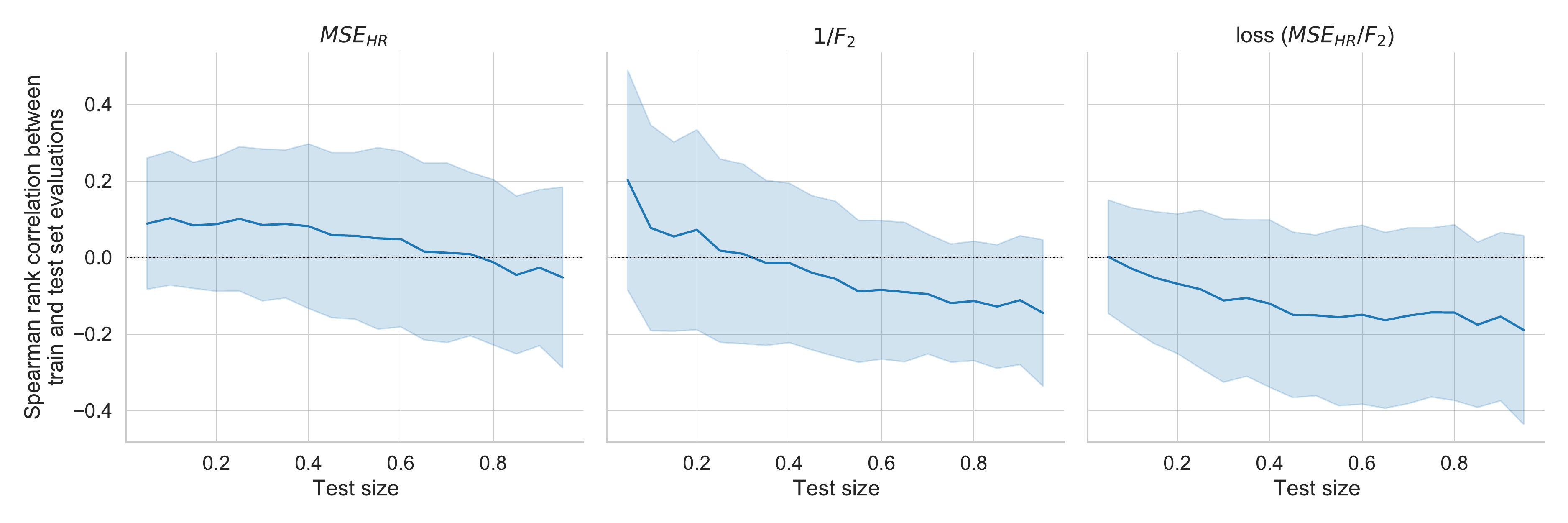}
    \caption{Extent to which different data splits impact ability to infer test set performance from the train set performance. Expected Spearman rank-order correlation coefficients between train and test set evaluations, as data sets vary in the fraction of events assigned to both (shown in the $x$-axis). Correlations measured in the \MSE\ and $1/F_2$ metrics, as well as their product.}
    \label{fig:comp_sim__different_splits__correlation}

    \vspace*{\floatsep}

    \centering
    \includegraphics[width=\textwidth]{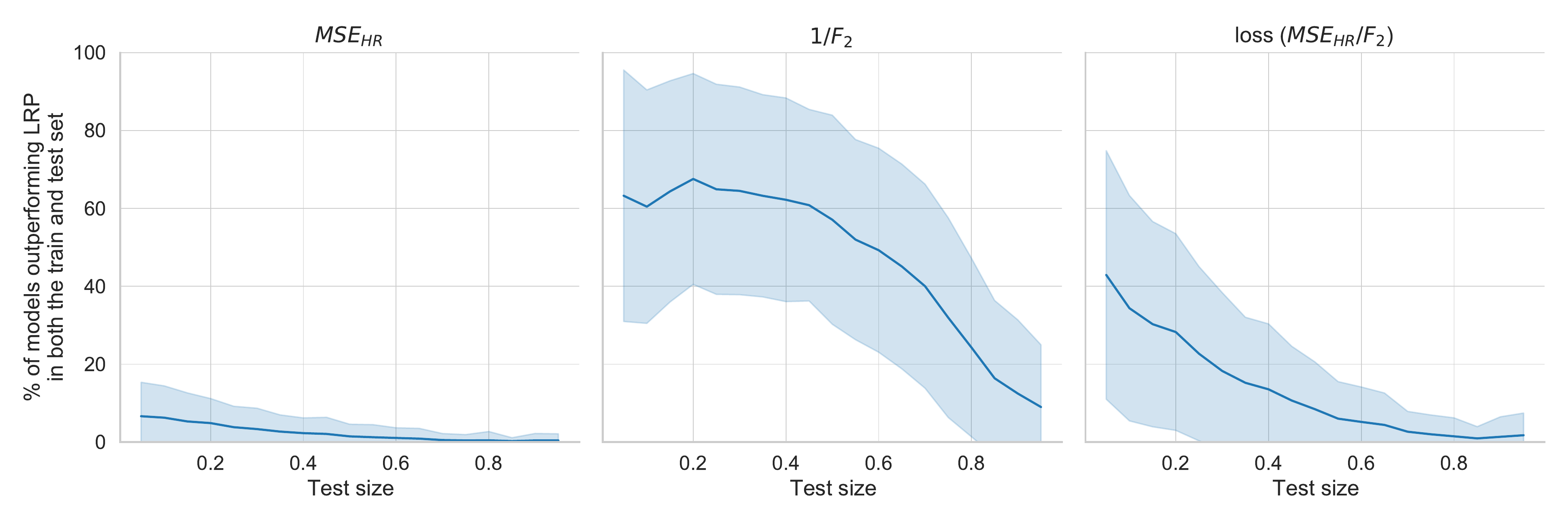}
    \caption{Expected percentage of machine learning models that will outperform the LRP baseline in both the train and test set, as data sets vary in the fraction of events assigned to both (shown in the $x$-axis). Performance measured in the \MSE\ and $1/F_2$ metrics, as well as their product.}
    \label{fig:comp_sim__different_splits__outperforming_LRP}
\end{figure*}

\begin{figure*}[t]
    \begin{minipage}[t]{.48\textwidth}
    \centering
    \includegraphics[width=\linewidth]{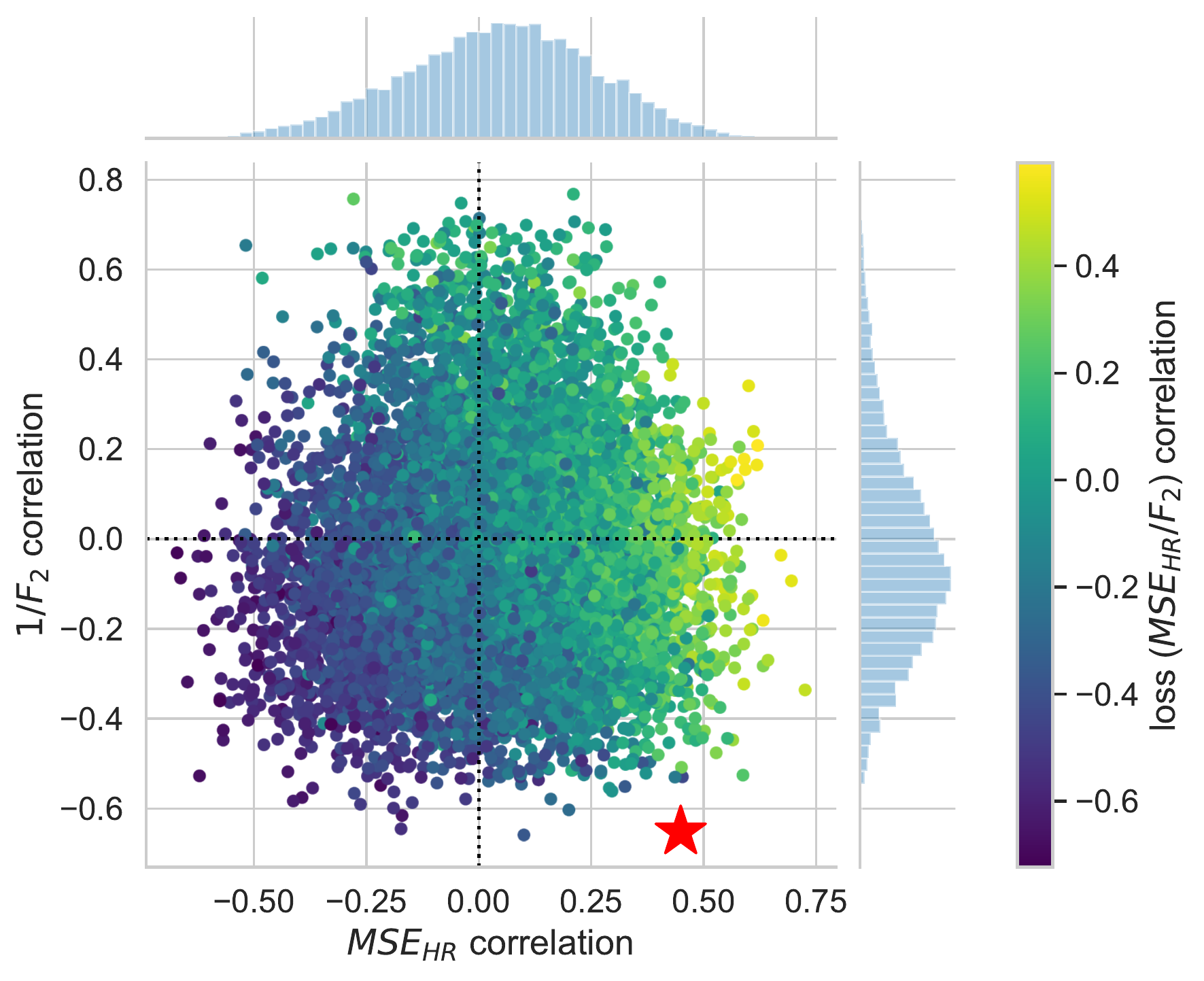}
    \caption{
    Extent to which different data splits impact ability to infer test set performance from the train set performance, as seen through simultaneous evaluations of regression and classification metrics.
    Spearman rank-order correlation coefficients between train and test set evaluations of the three performance metrics, in 10000 different data splits, using different test size fractions. Red star corresponds to the official competition's data split (Fig.~\ref{fig:comp_sim__official_split}).}
    \label{fig:comp_sim__different_splits__correlation_scatter}
    \end{minipage}
    \hspace*{\fill}
    \begin{minipage}[t]{.48\textwidth}
    \centering
    \includegraphics[width=\linewidth]{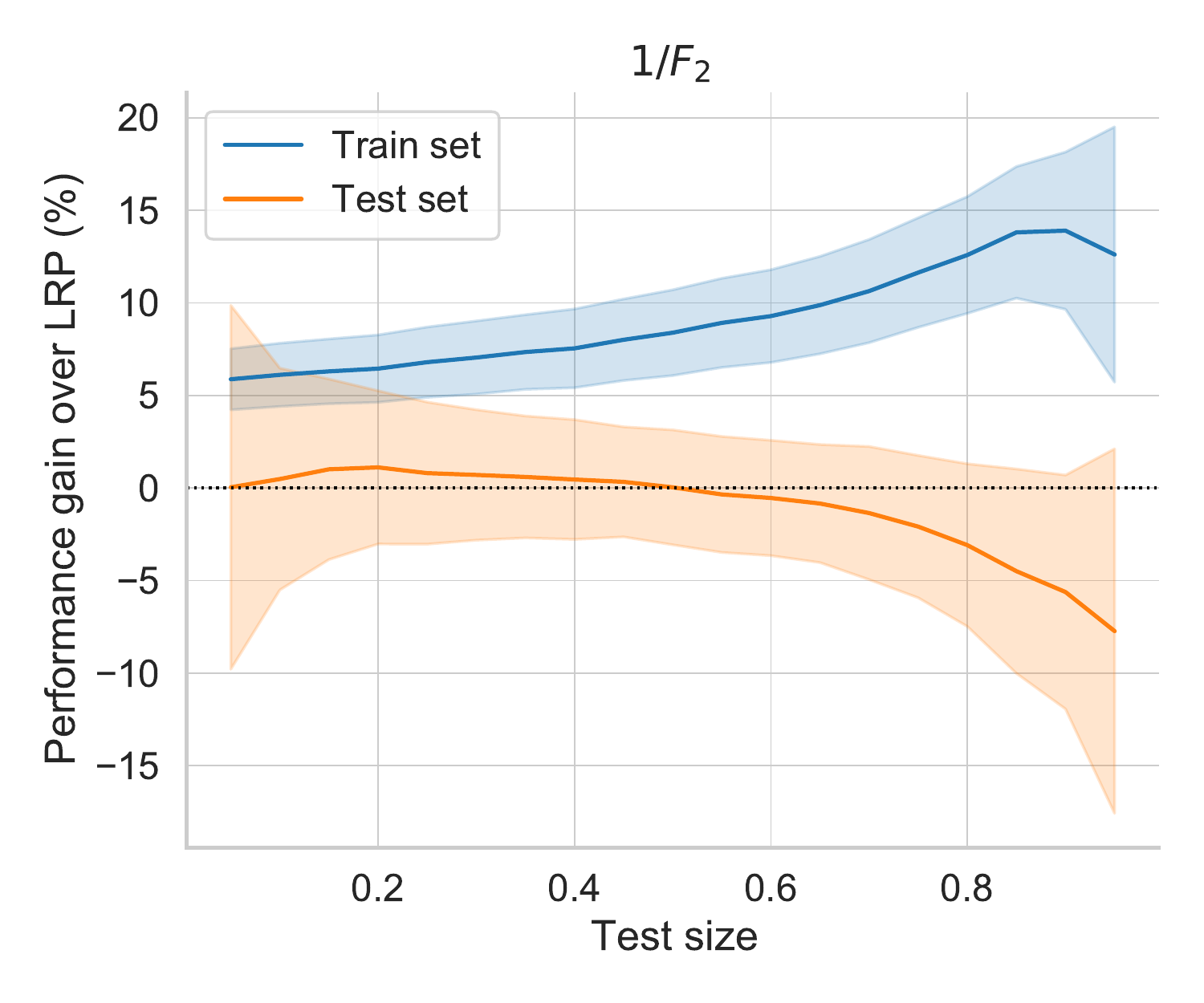}
    \caption{Expected $1/F_2$ performance gain ($\%$) over the LRP baseline as data splits vary in the fraction of events assigned to either the train or test set.}
    \label{fig:comp_sim__different_splits__1F2_perf_gain}
    \end{minipage}

\vspace*{\floatsep}

    \centering
    \includegraphics[width=\textwidth,viewport = 23 24 1253 304,clip]{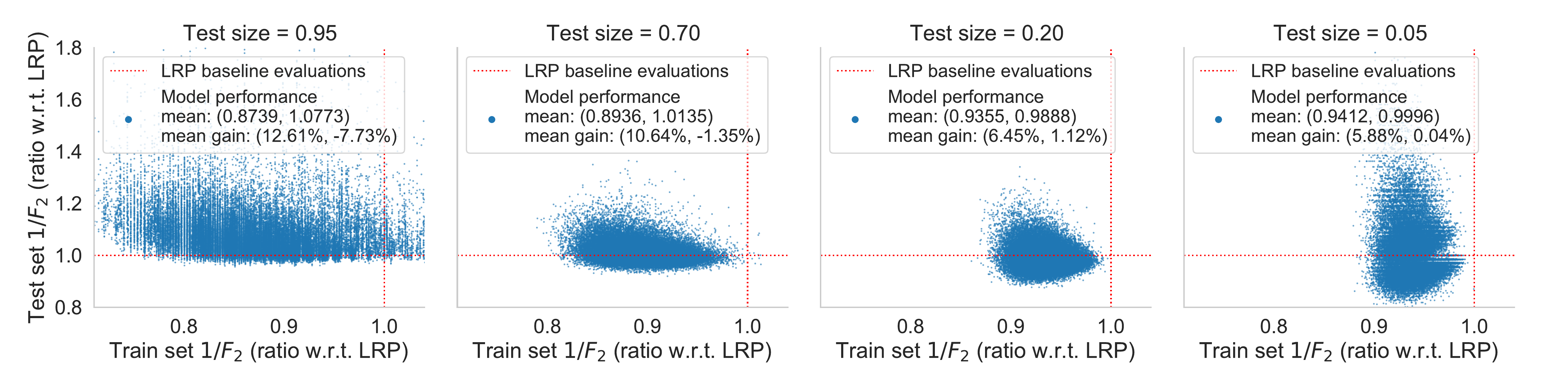}
    \caption{Variation in machine learning models' train and test set performance ($1/F_2$) as a function of data availability. Performance normalized with respect to the LRP baseline's performance in the same datasets.}
    \label{fig:comp_sim__different_splits__1F2_scatter}

\end{figure*}

The experiment's results can be seen in Figures~\ref{fig:comp_sim__different_splits__correlation}--\ref{fig:comp_sim__different_splits__1F2_scatter}.
Figure~\ref{fig:comp_sim__different_splits__correlation} shows statistics on the Spearman rank-order correlation coefficients between model evaluations in the train and test sets, per evaluation metric.
Positive Spearman correlation signals an ability to use the metric for model selection. The better a model is on the train set, the better we will expect it to be on the test set’s unseen events. Negative correlation is a sign of overfitting, or inability to generalize beyond the train set data.
Figure~\ref{fig:comp_sim__different_splits__outperforming_LRP} complements the analysis in Figure~\ref{fig:comp_sim__different_splits__correlation}, by showing statistics on the percentage of models, per simulated competition, that managed to outperform the LRP baseline in their respective train and test sets.
The curves show mean performance as a function of the test size, and the shaded areas represent the region within 1 standard deviation.
Figure~\ref{fig:comp_sim__different_splits__correlation_scatter} also shows correlations between train and test set evaluations, but now matches \MSE\ correlations to $1/F_2$ correlations. This way, it is possible to get an overview of how the same data split impacts models’ capabilities to learn generalizable knowledge simultaneously with respect to regression and classification goals. The red star places the Kelvins competition’s unstratified (with respect to high-risk events) data split in the context of 10000 stratified splits, showing just how much of an outlier it turned out to be.

The first conclusion to be drawn from these figures is that the aggregated loss metric, $\textrm{MSE}_{\textrm{HR}}/F_2$, is decidedly not informative in terms of identifying models that are likely to generalize. It takes two metrics that by themselves display low correlation between train and test set, and aggregates them into a single value which, as a result, is even less correlated.
%
%
Furthermore, we see in Figure~\ref{fig:comp_sim__different_splits__correlation_scatter} that the highest loss correlations tend to occur when \MSE\ is itself highly correlated.
In fact, \MSE\ is of the three metrics the one that tends to display a higher rank correlation. However, as seen in Figure~\ref{fig:comp_sim__different_splits__outperforming_LRP}, few models ever outperform the \MSE\ obtained by the LRP baseline, on both sets. This points to an identical situation to the one seen in Figure~\ref{fig:comp_sim__official_split}, where we have a high positive correlation, but models are not particularly successful. It is difficult to predict the actual final risk value, and so, the more our predictions move away from the high-risk events' most recent risk estimate (the only events scored by this metric), the worse we are likely to perform, both in the train and test sets.
Nonetheless, as seen in the $1/F_2$ plots of Figures~\ref{fig:comp_sim__different_splits__correlation}--\ref{fig:comp_sim__different_splits__outperforming_LRP}, even if the predicted final risk values are not accurate, those perturbations do move us across the -6 risk threshold in ways that result in improved capability to forecast the final risk class. With a test size of 0.2, 67.55\% of the trained models outperform the LRP baseline on both their train and test sets. This stands in stark contrast to Figure~\ref{fig:comp_sim__official_split}, where only 0.04\% of the models (2 out of 5000) outperformed the LRP baseline for the $1/F_2$ metric.

By normalizing models' $1/F_2$ evaluations with respect to the LRP baseline' $1/F_2$ values, performance becomes comparable across different data splits.
Figure~\ref{fig:comp_sim__different_splits__1F2_perf_gain} shows the mean and standard deviation of models’ percentage gains in performance over the LRP baseline, in the train and test sets. Statistics are calculated across all models trained over all data splits that used the same test size setting.
Figure~\ref{fig:comp_sim__different_splits__1F2_scatter} shows models’ $1/F_2$ evaluations in the train and test sets, normalized against their respective LRP $1/F_2$ baseline, for selected test size settings. Over 50000 machine learning models are shown in each subplot, trained over 500 data splits with that test size setting, on average.

At one end, with a test size of 0.95, train sets have merely 523.0 events to learn from (10.8 of which are of high risk). With insufficient data to learn from, models quickly overfit, and fail to learn generalizable rules. This can be seen by a mean gain in performance of 12.61\% on the train set, but a 7.73\% mean loss in performance on the test set, both with respect to the LRP baseline.
As we increase the amount of data available for training models, train set performance decreases (more data patterns to learn from, harder to incorporate individual events’ idiosyncrasies into the model), but test performance increases.
At the other end, with a test size of 0.05, most data is available for training, but the small test set is no longer representative (523.0 events to evaluate models on, 10.8 of high risk). Depending on the ``predictability'' of events that make it into the test set, we either obtain very high or very low performance: mean gain of 0.04\% on the test set, with a standard deviation of 9.83\%.
The optimal trade-off lies here at a test size of 0.2: we see a 6.45\% gain in train set performance over the LRP baseline, that transfers to a 1.12\% gain in the test set.
It is interesting to observe that we have experimentally converged on this as being the ideal setting, given the 80/20 split is a common rule of thumb among data scientists, inspired by the Pareto principle.
%
%
%
%

To establish a machine learning performance baseline, we now turn directly to the $F_2$ score, rather than its inverse (see discussion in Section~\ref{sec:metric}).
$F_2$, which ranges in $[0,1]$, is the harmonic mean of precision and recall, where recall (ability to identify all high-risk events) is valued 2 times higher than precision (ability to ensure all events predicted as being of high-risk indeed are).
A Monte Carlo cross-validation with a test size of 0.2, and 505 stratified random data splits evaluates the LRP baseline (the direct use of an event’s latest CDM’s risk value as prediction) to a mean $F_2$ score of 0.59967 over the test set (standard deviation: 0.04391).
Over the same data splits, a LightGBM regressor, acting over the same CDMs’ raw values (see data and model configurations in Section~\ref{sec:gen:official_split}), evaluates to a mean $F_2$ score of 0.60732 over the test set (standard deviation: 0.04895; statistics over 50500 trained models). So, a gain of 1.2743\% over the LRP baseline
\footnote{For comparison, cross-validation of team sesc's method (Section~\ref{sec:sesc}), over the same 505 data splits with test size of 0.2, evaluates to a mean $F_2$ score of 0.51563 over the test set. A performance loss of 14\% with respect to the LRP baseline.}.
The difference in performance between both approaches is statistically significant:
a paired Student’s t-test rejects the null hypothesis of equal averages (t-statistic: -10.23, two-sided p-value: $1.83 \times 10^{-22}$; per data split, LRP $F_2$ score is paired to mean LightGBM model $F_2$ score, to ensure independence across pairs).

This is the strongest evidence yet that machine learning models can indeed learn generalizable knowledge in this problem.
In a domain that safeguards assets valued in the millions, a 1\% gain in risk classification can already be transformative. Furthermore, this would be a 1\% gain on top of approaches for risk determination with decades of work behind them.
Note that these results, obtained using a classification metric, are achieved through a regression modelling approach. Furthermore, one with intentionally limited modelling capability, trained over a basic data preparation process, and evaluated under adverse conditions (due to imbalances in mission id and chaser object type). It is thus expected that it will be possible to significantly surpass these performance levels with more extensive work in data preparation and modelling.

\subsection{Feature relevance}

\begin{table*}[t]
\fontsize{7.75}{7.75}\selectfont
\centering
\caption{Feature relevance estimates. In prediction of near-term changes in risk, percentage of the reduction in error attributable to the feature.}
\label{tab:feature_relevance}
\begin{tabular}{lrrrll}
\toprule
                   \textbf{Feature} &  \textbf{Rank} &   \textbf{Mean} &  \textbf{Std. dev.} &                                                                   \textbf{Description} &  \textbf{Unit} \\
\midrule
                        risk  &   1 &  29.275 &  9.557 &                       self-computed value at the epoch of each CDM [base 10 log] &       \\
          max\_risk\_scaling  &   2 &  22.544 &  8.979 &                     scaling factor used to compute maximum collision probability &       \\
       mahalanobis\_distance  &   3 &   3.261 &  1.675 &   unitless miss distance rescaled using combined covariance of target and chaser &       \\
                 c\_sigma\_t  &   4 &   3.000 &  1.715 &         covariance; transverse (along-track) position standard deviation (sigma) &     m \\
         max\_risk\_estimate  &   5 &   2.624 &  1.367 &            maximum collision probability obtained by scaling combined covariance &       \\
              c\_sigma\_rdot  &   6 &   2.191 &  1.369 &                           covariance; radial velocity standard deviation (sigma) &   m/s \\
              miss\_distance  &   7 &   2.089 &  1.112 &                                relative position between chaser \& target at TCA &     m \\
c\_position\_covariance\_det  &   8 &   1.778 &  1.066 &         determinant of covariance ({\raise.17ex\hbox{$\scriptstyle\sim$}}volume) &       \\
                 c\_sigma\_n  &   9 &   1.312 &  0.625 &                    covariance; (cross-track) position standard deviation (sigma) &     m \\
               time\_to\_tca  &  10 &   1.236 &  0.517 &            time interval between CDM creation and time-of-closest approach (TCA) &  days \\
                 c\_sigma\_r  &  11 &   1.177 &  0.739 &                           covariance; radial position standard deviation (sigma) &     m \\
                c\_obs\_used  &  12 &   1.164 &  0.554 &                   number of observations used for orbit determination  (per CDM) &       \\
              c\_sigma\_ndot  &  13 &   0.964 &  0.437 &             covariance; normal (cross-track) velocity standard deviation (sigma) &   m/s \\
       relative\_position\_n  &  14 &   0.954 &  0.754 &                 relative position between chaser \& target: normal (cross-track) &     m \\
    c\_recommended\_od\_span  &  15 &   0.945 &  0.423 &                    recommended length of update interval for orbit determination &  days \\
       relative\_position\_r  &  16 &   0.835 &  0.440 &                               relative position between chaser \& target: radial &     m \\
                     c\_sedr  &  17 &   0.779 &  0.486 &                                                          energy dissipation rate &  W/kg \\
                         SSN  &  18 &   0.773 &  0.372 &                                                              Wolf sunspot number &       \\
                 c\_crdot\_t  &  19 &   0.718 &  0.468 &  covariance; correlation of radial velocity vs transverse (along-track) position &       \\
             relative\_speed  &  20 &   0.699 &  0.400 &                                   relative speed between chaser \& target at TCA &   m/s \\
\bottomrule
\end{tabular}
\end{table*}

The experiment described in the previous section provides us with a basis over which to quantify feature relevance, that is independent from the specifics of any given data split, or the decision-making of any individual model. 
%
We provide that information here, to shed light on what signal machine learning models use in order to arrive at their predictions, and to direct future work towards the more important features to train models on.

Of the 1 million models trained in the previous section's experiment, 47.75\%, from across different test size settings, managed to surpass the $1/F_2$ LRP baseline on both their train and test sets. We selected all those models, and had LightGBM quantify their ``gain'' feature relevance. This process measures not how often a feature is used across a model's decision trees, but the gains in loss it enables when that feature is used (loss here refers to the objective function optimized by the algorithm while building the model, not to the competition's $\textrm{MSE}_{\textrm{HR}}/F_2$ scoring function). Per model, relevance values are normalized over features' total gains, and converted to percentages. Values are then aggregated through weighted statistics across the selected models, resulting in the relevance assessments seen in Table~\ref{tab:feature_relevance} (only the top 20 features are shown, out of the 100 used). Models' fractional gains in performance over the test sets' LRP $1/F_2$ baseline are used as weights.

The LRP is a strong predictor, as previously seen. However, relevance measurements show that in the machine learning models, features directly tied to risk (\textit{risk}, \textit{max\_risk\_scaling}, \textit{max\_risk\_estimate}) together account for only half (54.44\%) of models' gains in loss. Models are making wide use of the information available to them, with the top 20 features in Table~\ref{tab:feature_relevance} accounting for 78.32\% of the gains in loss, and only 2 of the 100 features having a relevance of 0.0. 

A set of 40 features have values for both the ``target'' (the ESA satellite -- prefix \textit{t\_}), and ``chaser'' we want to avoid (space debris/object -- prefix \textit{c\_}), for a total of 80 of the 100 features. Note the absence of ``target'' features in Table~\ref{tab:feature_relevance}. The relevance of ``target'' features sum to a total of 9.41\%, while ``chaser'' features sum to 23.49\%.
If models were to tie themselves too much to properties of the ``target'', they would be learning mission-specific rules. Instead, we see a greater reliance on properties of the ``chaser'', and in features with relative values, thus enabling better generalization across missions.

The mean relevance estimates are very stable. The unweighted aggregation of normalized relevance values in the remaining 52.25\% of trained models not included in the selection above shows a total of 10.51\% absolute difference across features.
The higher performing models from which the statistics in Table~\ref{tab:feature_relevance} were drawn, display by comparison a greater reliance in \textit{risk} and \textit{max\_risk\_scaling} (+4.62\%).
\textit{SSN}, the Wolf sunspot number, at a rank of 18, is one of the most relevant features. It is also one of the features with greater increase with respect to to alternate ranking, climbing 3 positions, and increasing relevance by 0.07\%.

Note that models under consideration take CDMs' raw values as input. It is possible that with some data engineering, the attributes presented in Table~\ref{tab:feature_relevance} may follow a different ranking. Indeed, the information content of the new features with respect to the prediction target may become clearer to identify and use by the machine learning algorithms. Note also that correlated features might be splitting relevance values between them, and thus appear lower in the ranking.


\section{Lessons Learned}
\label{sec:lessons_learned}
Some valuable lessons were learned during the course of this competition, both in the organization and the design of the competition itself and in the scientific findings. These lessons are shared in the hope they may serve to improve future forecasting competitions in the complex domain of Spacecraft Collision Avoidance, as well as document what has been proved to be achievable using machine learning.

\begin{itemize}
    \item \textit{Competition metric}: overall the metric described in Eq.~(\ref{eq:metric}) revealed to be a troublesome choice as it resulted in an insufficient correlation among the test and training set. The $F_2$ metric alone appears to be a superior choice in this respect. Furthermore, the score clipping was not applied in the original metric definition and had to be found and applied by the various teams independently. Including such clipping already in the metric definition would have helped to uniform scores, discussions and make the various teams focus on more important aspects of the competition. 
    \item \textit{Training and test sets}: most teams, including the highest ranking ones, found that models leading to good scores on the training set did not generalize well to the test set. 
    While a detailed description of how the test and training sets were assembled was available to all participants, it did not seem to have helped the teams in processing the training set as to produce one that would correlate to the test set. As a consequence, many teams attempted to learn directly from the test set by probing the submission system of the web portal. This is unfortunate as the choice of putting a larger proportion of high-risk events in the test set was made in the hope that it would make probing the high-risk events (which affect the metric the most) more difficult. A deeper understanding on the consequences of the data split chosen on the competition dynamics was likely lacking and led to an unfortunate choice of the final split. 
    \item \textit{Baselines}: baseline solutions are useful to participants to get a feeling of what the organizers would consider as good results. The CRP baseline that was made available prior to the competition start, in this sense, failed as it turned out to be easily improved by the use of a large number of scientifically uninteresting solutions. 
    The LRP baseline, a slight variant to the widely known naive forecasting model was not disclosed and would have been a more competitive and natural choice. Leaving to the teams the task to find such a naive forecasting approach had the only effect to shift the competition focus from more interesting work.     
    \item \textit{Mission types}: the dataset released included CDMs that referred to different missions with heterogeneous characteristics. As a first step, it is probably worth to investigate the collision risk forecasting for a single satellite, to later study how well and if such a model can generalize to other missions. In addition, missions should be proportionally represented in the training and test sets, for both low-risk and high-risk events.
    \item \textit{Machine Learning}: variants to the naive forecast revealed to be surprisingly strong models for collision risk forecasting, hinting that CDMs time series might be close to following the Markov property, similarly to the time series encountered in economic and financial modeling~\cite{chen2012testing}. Due to the above observation, machine learning comes into a setting where it is difficult to extract signal that can be used to improve upon those forecasts. This result in a complex learning problem, and consequently, in a greater sensitivity to the design and evaluation setup. However, we  demonstrated, for the first time, the usefulness of machine learning for this problem domain. Finally, focusing on propagating the associated covariance (see Figure \ref{fig:cov_high_risk}) over time may offer a way to outperform solutions revolving around naive forecasting. A preliminary attempt in this direction can be found in \cite{Metz:2020}. 
    
\end{itemize}

\section{Conclusions}
\label{sec:conclusions}
The Spacecraft Collision Avoidance Challenge allowed, for the first time, to study the use of machine learning methods in the domain of spacecraft collision avoidance thanks to the public release of a unique dataset collected by ESA's Space Debris Office over the course of more than four years of operations. 
Several challenges, mostly deriving from the unavoidable unbalanced nature of the dataset, had to be accounted for to release the dataset in the form of a competition and limited the use of automated, off-the-shelves machine learning pipelines. 
Nevertheless the competition results and further experiments here presented clearly showed us two things. On one hand, naive forecasting models have surprisingly good performances and thus are established as an unavoidable benchmark for any future work in this area and, on the other hand, machine learning models are able to improve upon such a benchmark hinting at the possibility of using machine learning to improve the decision making process in collision avoidance systems. 


%

\section*{Acknowledgment}
ESA would like to thank the US Space Surveillance Network for the provision of surveillance data supporting safe operations of ESA’s spacecraft. In addition, we are grateful to the agreement that allowed to publicly release the dataset for the purpose of the competition.

The authors would like to thank all the scientists that participated to the Spacecraft Collision Avoidance Challenge and that dedicated their time and knowledge to what is an important element of ESA's operated satellites.
In particular we would like to acknowledge all members of the team \emph{sesc} whose methodology is shortly described in this paper: Steffen Limmer, Sebastian Schmitt, Viktor Losing, Sven Rebhan and Nils Einecke. 



\end{document}